\documentclass[a4paper,twocolumn]{article}
\usepackage[letterpaper,margin=1.5cm]{geometry}
\usepackage{multicol}
\usepackage{graphicx}
\usepackage{amsmath, amssymb,mathtools}
\usepackage{amsfonts}
\usepackage{textcomp}
  \usepackage[caption=false,font=footnotesize]{subfig}
\usepackage{soul}
\usepackage{booktabs, algorithm, algorithmic, multirow, hyperref}
\usepackage[table]{xcolor}
\usepackage[utf8]{inputenc}
\usepackage[english]{babel}
\usepackage{csquotes}
\usepackage[hyperref,
doi,
backend=biber,
style=ieee
]{biblatex}

\bibliography{paper.bib}

\AtEveryBibitem{
 \clearlist{address}
 \clearfield{date}
 \clearfield{eprint}
 \clearfield{isbn}
 \clearfield{issn}
 \clearfield{url}
 \clearlist{location}
 \clearfield{month}
 \clearfield{series}
 \clearfield{pages}
 \clearfield{volume}
 \clearfield{doi}
 \clearfield{number}
 \clearfield{booktitle}

 \ifentrytype{book}{}{
  \clearlist{publisher}
  \clearname{editor}
 }
}
\usepackage{authblk}
\newcommand{\diff}[2] { \frac{d#1}{d#2}}
\newcommand{\ddiff}[2] { \frac{{d}^2#1}{d{#2}^2}}
\newcommand{\loss}[0] {\mathcal{L}}
\newcommand{\s}[1] {\textsuperscript{#1}}
\newcommand{\norm}[1]{\left\lVert#1\right\rVert}
\newcommand{\abs}[1]{\left\lvert#1\right\rvert}
\newtheorem{finding}{Finding}

\usepackage{array}
\newcolumntype{L}[1]{>{\raggedright\let\newline\\\arraybackslash\hspace{0pt}}m{#1}}
\newcolumntype{C}[1]{>{\centering\let\newline\\\arraybackslash\hspace{0pt}}m{#1}}
\newcolumntype{R}[1]{>{\raggedleft\let\newline\\\arraybackslash\hspace{0pt}}m{#1}}

\begin{document}

\title{Taxonomy of Saliency Metrics for Channel Pruning}

\author{\uppercase{Kaveena Persand},
\uppercase{Andrew Anderson, and David Gregg}}
\affil{Authors are with the School of Computer Science and Statistics, Trinity
College Dublin, Ireland e-mail: persandk@tcd.ie, aanderso@tcd.ie,
david.gregg@cs.tcd.ie}

\date{}

\markboth
{Persand \MakeLowercase{\textit{et al.}}: Taxonomy of Saliency Metrics for Channel Pruning}
{Persand \MakeLowercase{\textit{et al.}}: Taxonomy of Saliency Metrics for Channel Pruning}

\begin{@twocolumnfalse}
\maketitle
\begin{abstract}

Pruning unimportant parameters can allow deep neural networks (DNNs)
to reduce their heavy computation and memory requirements. A
\textit{saliency metric} estimates which parameters can be safely
pruned with little impact on the classification performance of the
DNN. Many saliency metrics have been proposed, each within the context
of a wider pruning algorithm.
The result is that it is difficult to separate the effectiveness of
the saliency metric from the wider pruning algorithm that surrounds
it. Similar-looking saliency metrics can yield very different results
because of apparently minor design choices. 
We propose a taxonomy of saliency metrics based on four
mostly-orthogonal principal components. We show that a broad range of
metrics from the pruning literature can be grouped according to these
components.  Our taxonomy not only serves as a guide to prior work,
but allows us to construct new saliency metrics by exploring novel
combinations of our taxonomic components. We perform an in-depth
experimental investigation of more than 300 saliency metrics. Our
results provide decisive answers to open research questions, and
demonstrate the importance of reduction and scaling when pruning
groups of weights.  We find that some of our constructed metrics can
outperform the best existing state-of-the-art metrics for convolutional neural
network channel pruning.

\end{abstract}
\end{@twocolumnfalse}

\section{Introduction}


Deep neural networks (DNNs) now offer human-level or
greater accuracy for many decision problems
~\autocite{DBLP:conf/ppopp/HestnessAD19,DBLP:conf/iccv/HeZRS15,DBLP:conf/cvpr/TaigmanYRW14}.
However, DNNs can require enormous computation and memory resources,
which may be unavailable in mobile and embedded devices where audio,
image and other data often originate. Transferring this data
off-device to the cloud for DNN processing creates many problems with
latency of response, energy, legal and privacy issues.



One way to reduce the resource requirements of trained DNNs is to \emph{prune}
unused parameters, or more concretely, to replace some of the values in weight
tensors by zero. An ideal pruning algorithm would remove the maximal number of
weights from a network while maintaining or improving accuracy.
While there is a huge variety of pruning schemes in the literature, the vast
majority have, at their heart, a \emph{saliency metric}. A saliency metric is
used to answer a fundamental question in pruning: which weight, or
set of weights, when removed, will likely cause the least damage to the network
predictions? Since the saliency metric is typically presented within the
context of a larger pruning algorithm, it is often extremely difficult to
isolate the effect of the saliency metric from other design choices.

Pruning of DNN weight tensors can be performed at different levels of
granularity~\autocite{Dally,LearningStructuredSparsity}, from
individual weights~\autocite{LWC} to large sub-blocks. Individual
weights typically participate in the computation of many different
elements of the output feature map. Thus, at most levels of pruning
granularity, saliency metrics focus on the weight itself rather than
the output feature maps computed using the weight. However, for
convolutional neural networks, there is one level of pruning
granularity --- channel pruning --- where there is a direct
relationship between sub-blocks of the weight tensor and sub-blocks of
the output feature map. We therefore focus on channel pruning, which
allows us to compare metrics based on weights or ouput feature maps.
Pruning full channels also yields a \emph{dense} weight tensor which
can be used with existing highly-optimized DNN libraries.


\begin{figure}[t]
\begin{center}
\includegraphics[width=0.7\columnwidth]{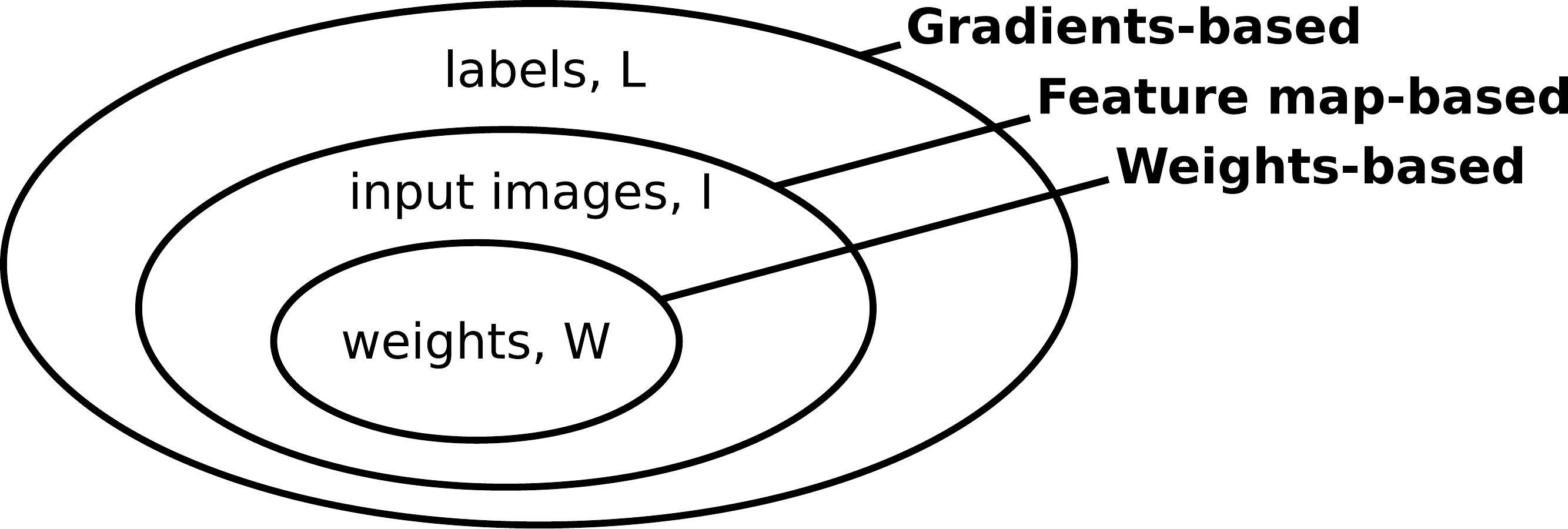}
\caption{Saliency metrics can be grouped into three categories depending
on the information that they use.}
\label{fig:information_taxonomy}
\end{center}
\end{figure}

\subsection*{Contributions}

In this paper, we study the impact of the choice of saliency metric with a
canonical channel pruning algorithm for CNNs. Although our empirical results
are for this specific context, there is a strong argument for the generality of
some of our findings, which we highlight in discussion. We make the
following specific contributions.



\begin{itemize}
\item We propose a taxonomy that classifies saliency metrics based
  on four mostly-orthogonal components.
\item We empirically evaluate 308 saliency metrics,
  including metrics from prior work and novel metrics derived from new
  combinations of taxonomic components.
\item We experimentally confirm a widely-acknowledged rule of thumb: gradient-based approaches as a class significantly
  outperform simpler weight-based methods.
\item We answer open research questions, such as whether the popular strategy of ignoring first-order terms in Taylor-expansions is safe in practice.
\item We find the non-obvious result that the choice of dimensionality reduction and parameter scaling method has a large impact on saliency metrics, and we propose an effective novel scaling method for channel pruning.
\item We show that good saliency metrics can be
  effective even without any subsequent fine-tuning/retraining, or greatly reduce the number
  of iterations required if retraining is used.
\end{itemize}

\section{Background}


A large number of pruning schemes have been proposed. Most pruning schemes
either modify the training process to gradually prune weights or have an
explicit discrete step that removes weights from the network.
As well as the mechanism used to drive weights to zero, pruning schemes also
determine at which stage of the training process pruning should apply.  While
early work focused on pruning fully trained
networks~\autocite{Mozer_Smolensky,Karnin90,Lecun, Hassibi}, recent work has
shown that various methods can be used to remove parameters from the network at
different stages of the training process~\autocite{Frankle_Carbin, SNIP}.

The pruning schemes can be categorized according to three dimensions: pruning
method, training strategy, and estimation criterion~\autocite{CNNPruningASurvey}. 
Saliency metrics or estimation criteria~\autocite{CNNPruningASurvey} are used
to estimate which weights can be pruned from the network.  Saliency metrics are
used whether pruning is incorporated into the training algorithm 
~\autocite{Reed, ADMM_Regulizer, GlobalSparseMomentum, GroupWiseBrainDamage,
GateDecorator, Sparse_ARM, Play_and_Prune, AutoBalance} or occurs in discrete
steps which sets weights to zero~\autocite{Hu, Dally, Molchanov, Lecun,
Hassibi, DSD, Yang, Frankle_Carbin, SNIP}.  The same saliency metric can be
used in very different pruning schemes. For example, the L1-norm of weights has
been used as saliency metric with simple pruning schemes~\autocite{LWC,Dally},
with probabilistic pruning~\autocite{StructuredProbabilisticPruning}, with
reinforcement learning~\autocite{AMC}, and even for pruning at
initialization~\autocite{Frankle_Carbin}.  Saliency metrics can also be used
when pruning in Winograd domain~\autocite{WinogradPruningDally,
WinogradPruningTang} and frequency domain~\autocite{FrequencyDomainDynamicPruning}.

Saliency metrics can be grouped depending the information used to
compute them.  Figure \ref{fig:information_taxonomy} groups saliency metrics
according to the information that they use.  Saliency metrics that use only the
weights have the advantage of having all required information readily
available. Data driven approaches require training data to make pruning
decisions.  Approaches that only use input images to make pruning decisions
require only forward passes of the network.  Approaches that use gradients
additionally require backward passes of the network to compute the loss with
respect to input labels and hence the gradients. The main differentiating
factor between these classes of approach in practice
(Figure~\ref{fig:information_taxonomy}) is the \emph{cost} associated with the
use of more information.

Saliency metrics can also be grouped according to how they identify least
important weights~\autocite{CNNPruningASurvey}.  Simple metrics like the
L1-norm of weight~\autocite{LWC} or APoZ~\autocite{Hu} assume that small
weights or feature maps with high frequencies of zeros are less important to
the network.  Taylor expansion-based metrics often approximate the change in
global~\autocite{Lecun, Hassibi, Molchanov, GlobalSparseMomentum} or
layer-wise~\autocite{Dong17} loss caused by pruning and remove weights that cause the
least change in loss.  ThiNet~\autocite{ThiNet2019}, He et
al.~\autocite{He2017}, and Lin et al.~\autocite{DiscrimationAwarePruning}
choose weights that lead to the least feature map reconstruction error.  Hur
and Kang~\autocite{HurKang} use the entropy of the weights to determine which weights
are least important.  While these saliency metrics are derived with different
assumptions, we can compare them by expressing them in a standard form.

\section{A Taxonomy of Saliency Metrics}

\begin{table}[h]
\begin{center}
\scriptsize
\begin{tabular}{m{0.99cm}C{2.8cm}C{1.45cm}C{1.25cm}}
 \textbf{Base} & \textbf{Pointwise Metric} & \textbf{Reduction} & \textbf{Scaling}\\
 \textbf{Input} & $f(x)$ & $R$ & $K$\\
\toprule
 $X=W$
& $x$
& $\sum\limits_{x \in \prescript{l}{}{X}_i} x $
& $ 1 $\\

 $X=A$
& $\diff{\loss}{x} $
& $\sum\limits_{x \in \prescript{l}{}{X}_i} \abs{ x }$
& $n(\prescript{l}{}{X}_i)$ \\

& $-x \frac{d\loss}{dx} $
& $ \abs{ \sum\limits_{x \in \prescript{l}{}{X}_i} x } $
& $\norm{\prescript{l}{}{\widetilde{S}}}_1 $ \\

& $-x \frac{d\loss}{dx} + \sum\limits_{y \notin \tilde{W}} \frac{xy}{2}\frac{d^2\loss}{dxdy}$
& $\sum\limits_{x \in \prescript{l}{}{X}_i} (x)^2 $
& $\norm{\prescript{l}{}{\widetilde{S}}}_2 $ \\

& $x - reconstruction(x)$
& $ \left ( \sum\limits_{x \in \prescript{l}{}{X}_i} x \right ) ^2 $
& $n(\mathcal{TC}(\prescript{l}{}{W}_i))$ \\

& $alternateBackprop(x)$
& $ \sqrt{ \sum\limits_{x \in \prescript{l}{}{X}_i} \left ( x \right ) ^2} $
&  \\
\midrule
\end{tabular}
\caption{A taxonomy of published channel saliency metrics. One component from each column is chosen to construct a channel saliency metric.}
\label{tab:taxonomy}
\end{center}
\end{table}


We describe existing saliency metrics using a taxonomy of four
principle components.  A fine-grain saliency metric is constructed
from a pointwise metric $F$ over the parameter set $X$. When we prune
larger groups of weights, such as entire channels, we need to combine
the saliency of individual weights into a metric for the entire
channel. We do this with a dimensionality reduction $R$ and
normalization $K$. Some examples of choices for $X$, $F$, $R$ and $K$
are given in Table \ref{tab:taxonomy}.

The general form of the saliency metric for an
arbitrary subset of parameters $X$ is given by Equation \ref{eqn:taxonomy}.

\begin{equation}
S = \frac{1}{K} \cdot \widetilde{S} \text{, with } \widetilde{S} = R \circ F(X)
\label{eqn:taxonomy}
\end{equation}

We introduce some notation to facilitate the description and comparison of
different saliency metrics. Consider a CNN with loss function $\loss$, and
trained parameters vector $W$. Let $\widetilde{W}$ represent the pruned
parameters vector and $n(\cdot)$ the cardinality of a tensor or vector, then
$n(W) > n(\widetilde{W})$.  

\noindent In our mathematical treatment, we consider this general case unless
stated otherwise. However, in our experimental evaluation, we are concerned with
parameter subsets $X$ corresponding to the parameters $\prescript{l}{}{W}_i$ of
a \emph{channel} of a convolutional layer. Thus, $\prescript{l}{}{S}_i$ denotes
the saliency of the $i^{th}$ channel of the $l^{th}$ convolution layer of the
CNN.


\subsection{Domain (choice of $X$)}


In our taxonomy, saliency metrics can be based upon the weights
themselves ($X=W$), or the values of output features maps that are computed
using the weights ($X=A$). There is often a close relationship between
the magnitude of a weight and the sensitivity of the DNN to pruning
the weight, so many saliency metrics use the weight as a key input.
However, in the specific case of pruning entire output channels, there is
a direct relationship with the output points corresponding to the
 $i^{th}$ channel of the $l^{th}$ convolution layer $\prescript{l}{}{C}_i$ (i.e. the feature map $\prescript{l}{}{A}_i$).
Removing all parameters contributing to an output feature map in a
convolutional layer results in the feature map becoming zero. When this
happens, all of the operations which are transitively used to compute the operation can also be pruned, resulting in large savings.

\begin{figure}[h]
\begin{center}
\includegraphics[width=0.9\columnwidth]{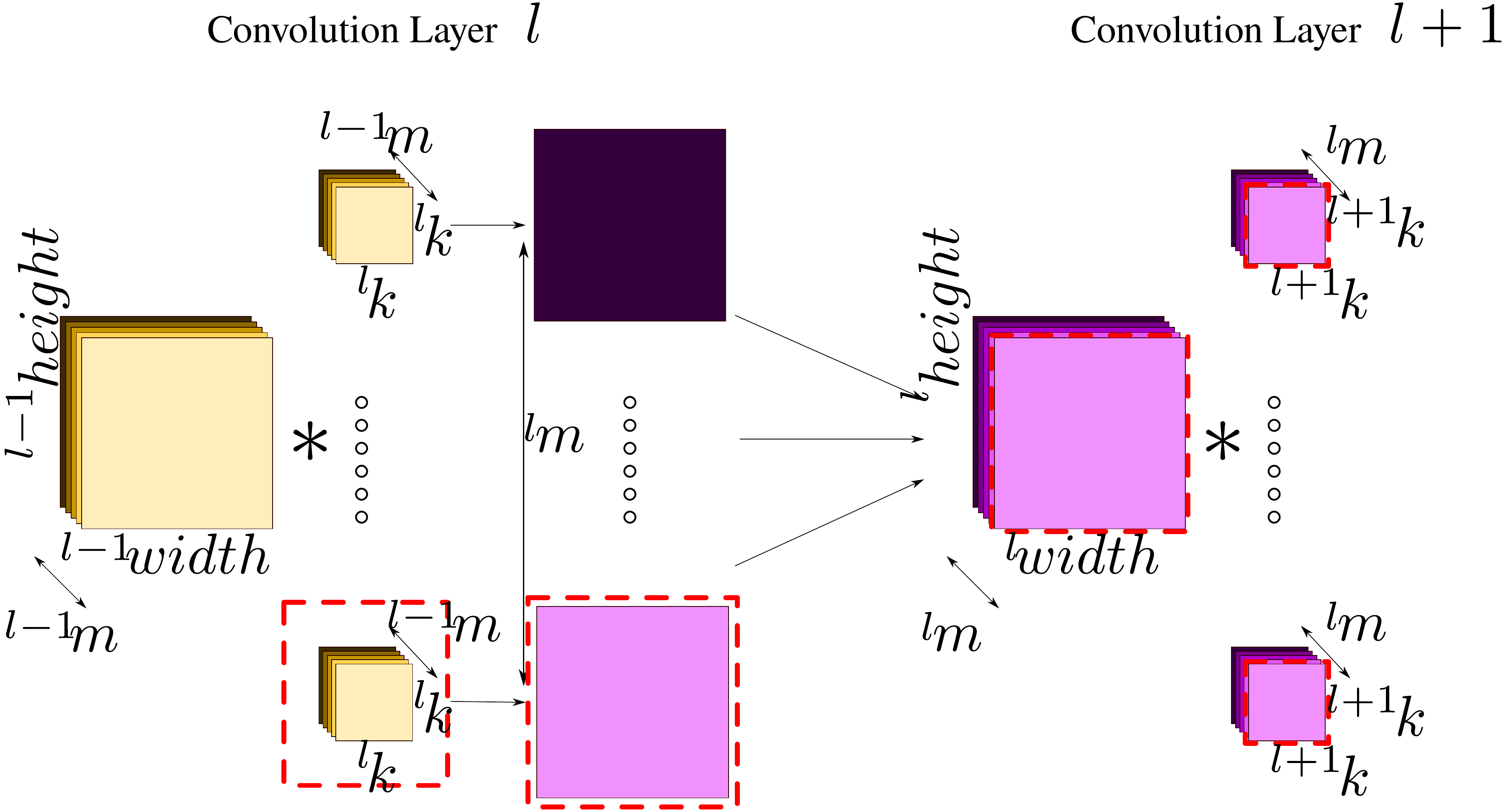}
\caption{Parameter/Feature map structure in a convolutional neural network.}
\label{fig:conv_pruning_figure}
\end{center}
\end{figure}

Hence, the saliency of a channel can be regarded as a function of outputs
~\autocite{Hu,Gaikwad,Anwar,Polyak_Wolf}, rather than parameters, i.e. in
Equation \ref{eqn:taxonomy} $X$ can be either the weights,
$\prescript{l}{}{W}_i$, or the output feature map, $\prescript{l}{}{A}_i$. The
relationship between the weights and feature maps of a channel is illustrated
in detail in Figure~\ref{fig:conv_pruning_figure}.

It should be noted that since channel pruning can be viewed as feature map
removal, feature selection metrics can also be used as saliency metrics for
pruning~\autocite{SCOP}.

When pruning at granularities finer than entire channels, the removal of output
points cannot be directly mapped to the removal of sets of weights.
Hence, it is clear that $X=W$ is the best choice. However, for channel pruning
it is difficult to choose definitively between output feature map or weights.
In fact, most published saliency metrics are presented either as functions of
weights or of output points, despite being applicable to both. Some metrics
~\autocite{Lecun,Hassibi} were originally defined using weights (as they were
used to prune individual weights), but derived metrics
~\autocite{Molchanov,Theis} use output feature maps instead.




To better illustrate the effect of the four orthogonal choices, we use an
example for channel pruning.  In Figure \ref{fig:taxonomy}, we show how to
compute the saliency of a convolution layer's channels using its weights.  This
corresponds to the case where $X=W$.

\subsection{Pointwise metric (choice of $F$)}

We denote $F(X)$ the tensor of pointwise saliency of all individual weights
or output points.  $F(X)$ is of the same shape as $X$, that is, either of the
shape of $W$ or $A$.  When pruning, it is common to look at either the saliency
of an individual element of the saliency vector or at a group of them.  To
facilitate this grouping, we introduce $\prescript{l}{}{F(X)}_i$ and $f(x)$.
$\prescript{l}{}{F(X)}_i$ is the tensor of saliency corresponding to the
$i^{th}$ channel of the $l^{th}$ layer.  $\prescript{l}{}{F(X)}_i$ is of the
same shape as $\prescript{l}{}{X}_i$.  Hence if $X=W$ or $X=A$, then
$\prescript{l}{}{F(X)}_i$ is of the shape $\prescript{l-1}{}{m} \times
\prescript{l}{}{k} \times \prescript{l}{}{k}$ or $\prescript{l}{}{height}
\times \prescript{l}{}{width}$ respectively.  $f(x)$, is used to denote the
saliency of a single weight or output point. If $x=\prescript{l}{}{W}_i[p,q,r]$
is an individual weight from $\prescript{l}{}{W}_i$, then
$f(x)=\prescript{l}{}{F(X)}_i[p, q, r]$.  Similarly when using the output
feature map instead of the weights, if $x=\prescript{l}{}{A}_i[p,q]$ is an
individual output point then $f(x)=\prescript{l}{}{F(X)}_i[p,q]$.

A common pointwise saliency function is the absolute magnitude function, i.e.
the saliency of an individual weight or output point is given directly by its
absolute value, hence $f(x) = \abs{x}$.

\begin{figure*}[]
\begin{center}
\includegraphics[width=1.0\textwidth]{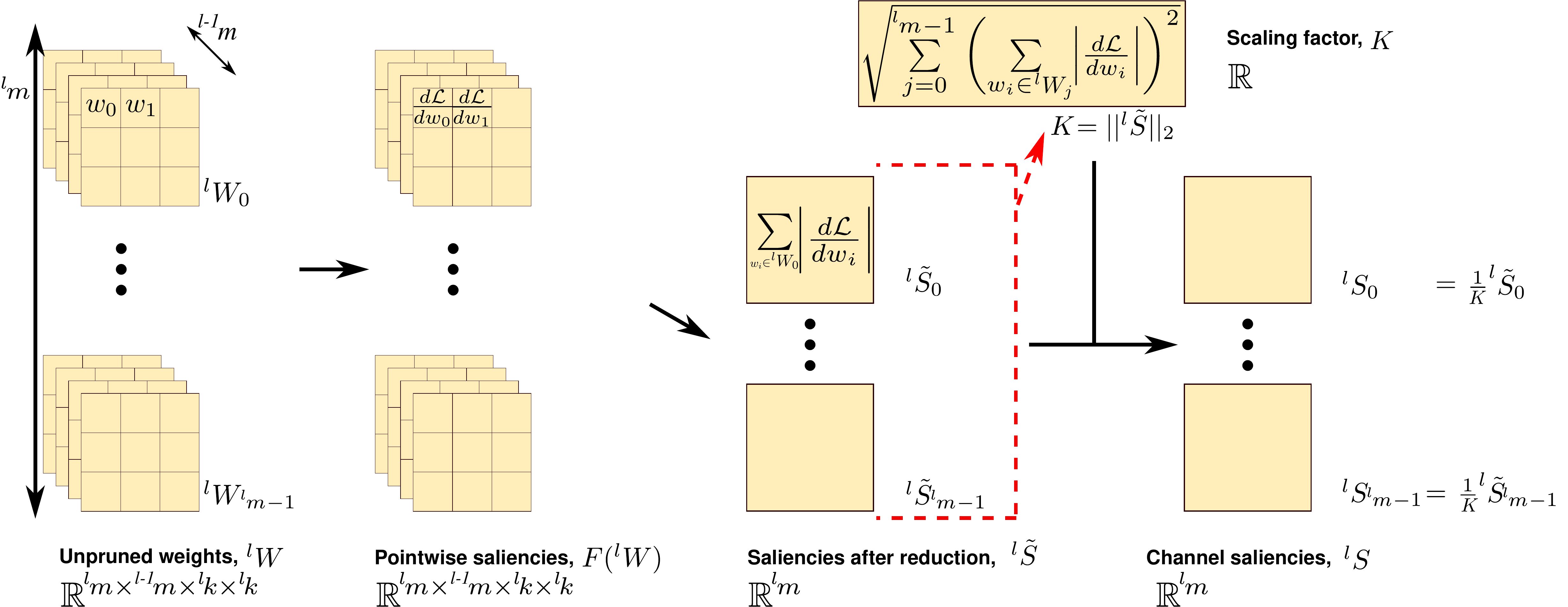}
\caption{Computing channel saliency using Equation \ref{eqn:taxonomy} using $X=W$, $f(x)=\diff{\loss}{x}$, $R(X) = \sum\limits_{x \in \prescript{l}{}{X}_i} \abs{x}$ and $K = \norm{\prescript{l}{}{\widetilde{S}}}_2$.}
\label{fig:taxonomy}
\end{center}
\end{figure*}

In Figure \ref{fig:taxonomy}, the gradient of an element is used as a saliency
metric, $F(X) = J $ or $f(x) = \diff{\loss}{x}$.  Applying $F$ to $W$ yields a
tensor containing the saliency of the individual weights.



\subsection{Dimensionality Reduction (choice of $R$)}

\noindent Once the pointwise saliency vector is obtained, a reduction  is used
to condense the tensor of pointwise saliency to a single value for the
pruning element.  In the case of channel pruning, $R$ reduces either a
$\prescript{l-1}{}{m} \times \prescript{l}{}{k} \times \prescript{l}{}{k}$
tensor or a $\prescript{l}{}{height} \times \prescript{l}{}{width}$ tensor into
a single value.

Any suitable vector norm could be used as a reduction, with the
L2-norm being a popular reduction method in the literature 
~\autocite{SoftFilterPruning, Gaikwad}.

In Figure \ref{fig:taxonomy} an L1-norm is used to reduce the
$\prescript{l}{}{m} \times \prescript{l}{}{k} \times \prescript{l}{}{k}$ tensor
of weight saliency into a vector containing the layer's channel saliency
,$\prescript{l}{}{\widetilde{S}}$, of dimension $\prescript{l}{}{m}$.




\subsection{Scaling (choice of $K$)}
\label{sec:scaling}

All other things being equal, the more parameters which can be pruned, the
lower the computational and memory costs of inference. Therefore, if two
channels have similar saliency, one should favour pruning the larger channel,
which is the Pareto-optimal choice considering the twin objectives of
minimizing accuracy loss and maximizing the number of parameters pruned.

To better integrate this cost function, a scaling coefficient, $K$ is typically
used. One solution is to scale the channel saliency $\prescript{l}{}{S}_i$
using the cardinality of the pointwise saliency vector. In other words, one
can also look at the average saliency of a group instead of the sum of the
saliency in the group.  For example, instead of using the sum of the
magnitudes or L1-norm of the weights, $\norm{\prescript{l}{}{W}_i}_1$, one can
use the average of the magnitudes, $\frac{1}{n(\prescript{l}{}{W}_i)}
\norm{\prescript{l}{}{W}_i}_1$. This nomalizes the result of the reduction, $R$,
so that channels with many weights are more likely to be pruned, leading to
greater overall sparsity. Another solution is to perform a
\emph{layer-wise} normalisation to scale the magnitudes of saliency across
layers.  Using a layer-wise L2-norm in the case of global pruning helps when
values for saliency in different layers have drastically different magnitudes
~\autocite{Molchanov}.

In Figure \ref{fig:taxonomy}, we use a layer-wise L2-norm as scaling factor.  We
use the L2-norm of the unscaled saliency of all the channels in the given 
convolution layer, $\prescript{l}{}{\widetilde{S}}$, to obtain the scaling
coefficient, $K$, of that layer.

\subsubsection{Proposed scaling method, $\mathcal{TC}$}

We propose to investigate a new scaling method, $\mathcal{TC}$.
$\mathcal{TC}(\prescript{l}{}{W}_i)$ is used to denote the entire set of
weights transitively removed when $\prescript{l}{}{W}_i$ is removed from the
network.  A simple example of $\mathcal{TC}(\prescript{l}{}{W}_i)$ is shown in
Figure \ref{fig:conv_pruning_figure}, where weights from the next convolution
layer are also removed when we remove an output channel from the network. Since
$\mathcal{TC}(\prescript{l}{}{W}_i)$ include $\prescript{l}{}{W}_i$,
$n(\mathcal{TC}(\prescript{l}{}{W}_i)) \geq n(\prescript{l}{}{W}_i)$.
When optimising for the maximum number of weights
removed for the least loss in accuracy, $\mathcal{TC}(\prescript{l}{}W_i)$ is
interesting because it takes into account all the weights removed for that
channel.

\subsection{Minibatches}

Data driven approaches which consider gradients or output points in the domain
$X$ often rely on a set of inputs to produce these values. Since
each input in the set will result in a potentially different set of
saliency values, the computation of the saliency over a minibatch is typically
done by combining the element-wise saliency values using a simple average across
the minibatch~\autocite{Theis,Molchanov,Gaikwad,Hu}.

Hence, the full saliency equation used in practice is given by Equation
\ref{eqn:taxonomy_batches} with $N$ the total number of images used in the
minibatch.

\begin{equation}
\prescript{l}{}{S}_i = \frac{1}{N}\sum\limits_{n = 0}^{N-1}\frac{1}{K} \cdot R \circ F(X_{n})
\label{eqn:taxonomy_batches}
\end{equation}

In some cases, a square root is applied to the resulting saliency metric
~\autocite{Gaikwad}.  This additional computation does not modify the ranking
of the channels and can thus be omitted for algorithms only concerned with
channel ranking.

\section{Classification of existing saliency metrics}

In this section, we classify popular saliency metrics using our taxonomy.  Table
\ref{tab:classification_channel_pruning} summarizes some channel
saliency metrics that have been used for pruning convolution channels.
We can obtain new saliency metrics by selecting different combinations of
of the four components in Table
\ref{tab:taxonomy}. Although some of these
saliency metrics resemble each other, their efficacy can vary.

This paper focuses on using saliency metrics specifically for channel
pruning. However, the taxonomy can also be used to classify saliency metrics
used for different granularities. For fine-grain pruning, i.e. pruning
individual weights, only the pointwise saliency is relevant.  For other
granularities of pruning, pointwise metrics are computed across the relevant
substructure in parameter tensors and then dimensionally reduced (e.g. with a
vector or tensor norm) to yield structural metrics. Table
\ref{tab:classification_fine_grain} shows the classification of a selection of
popular pointwise saliency metrics that have been used for fine-grain
pruning.

\begin{table}[h!]
\begin{center}
\caption{Published approaches for fine-grain pruning.}
\label{tab:classification_fine_grain}
\scriptsize
\begin{tabular}{lc}
\textbf{Method} & \textbf{Pointwise Measure, $f(x)$} \\
\toprule
Magnitude~\autocite{LWC, DSD, AMC, Frankle_Carbin, DynamicNetworkSurgery}
& $\abs{w}$\\
Optimal Brain Damage~\autocite{Lecun}
& $\ddiff{\loss}{w}$ \\
Optimal Brain Surgeon~\autocite{Hassibi}
& $\frac{w_i w_j}{2}\frac{d^2\loss}{dw_idw_j}$ \\
Gradient of mask~\autocite{Mozer_Smolensky}
& $- \diff{\loss}{m} $\\
Gradient of weights~\autocite{Karnin90}
& $- \diff{\loss}{w}$ \\
1st order Taylor expansion~\autocite{GlobalSparseMomentum}
& $ \abs{ \frac{d\loss}{dw}w }$ \\
\end{tabular}
\end{center}
\end{table}

\begin{table*}[h]
\begin{center}
\scriptsize
\begin{tabular}{llm{4.6cm}ll}
\textbf{Method} & \textbf{Base Input,} & \textbf{Pointwise Measure,} & \textbf{After Reduction,} & \textbf{Scaling,}\\
 & $X$ & $f(x)$ & $\prescript{l}{}{\widetilde{S}}_i$ & $K$ \\
\toprule
\textbf{Using only weights} & & & & \\
\midrule
L1-norm of weights~\autocite{Dally, Li2017, StructuredProbabilisticPruning}
& $W$
& $x$
& $\sum\limits_{x \in i\prescript{l}{}{X}_i} \abs{ f(x) } $
& $ 1 $ \\

L2-norm of weights~\autocite{GroupWiseBrainDamage, SoftFilterPruning}
& $W$
& $x$
& $\sum\limits_{x \in \prescript{l}{}{X}_i} f(x)^2$
& $ 1 $ \\

Min-weight~\autocite{Molchanov}
& $W$
& $x$
& $\sum\limits_{x \in \prescript{l}{}{X}_i} f(x)^2 $
& $\norm{ \prescript{l}{}{X}_i }_0$ \\

NISP~\autocite{NISP}
& $W$
& $alternateBackprop(x)$ with $alternateBackprop(x)=NISP(x)$ (Equation \ref{eqn:nispPoint})
& $\sum\limits_{x \in \prescript{l}{}{X}_i} f(x) $
& $1$ \\

Geometric median of weights~\autocite{GeometricMedian}
& $W$
& $x- reconstruction(x)$ with $reconstruction(x) = GM(x)$ (Equation \ref{eqn:GeometricMedianPoint})
& $\sum\limits_{x \in \prescript{l}{}{X}_i} f(x)^2$
& $1$ \\

\midrule
\textbf{Using weights and input images} & & & & \\
\midrule

Sum of feature map~\autocite{Anwar}
& $A$
& $x$
& $\sum\limits_{x \in \prescript{l}{}{X}_i}  f(x) $
& $ 1 $ \\

APoZ~\autocite{Hu}
& $A$
& $\begin{cases}
 1, \text{if } x > 0\\
 0, \text{else}
\end{cases}$
& $\sum\limits_{x \in \prescript{l}{}{X}_i} f(x)$
& $n( \prescript{l}{}{X}_i)$ \\

L2-norm of activations~\autocite{Gaikwad}
& $A$
& $x$
& $\sum\limits_{x \in \prescript{l}{}{X}_i} f(x)^2 $
& $1$ \\

\midrule
\textbf{Using weights, input images and labels} & & & & \\
\midrule

Fisher information using activations~\autocite{Theis, Turner}
& $A$
& $x\frac{d\loss}{dx}$
& $\left ( \sum\limits_{x \in \prescript{l}{}{X}_i} f(x) \right ) ^ 2 $
& $\frac{1}{2}$ \\

Fisher information using weights~\autocite{Molchanov2019}
& $W$
& $x\frac{d\loss}{dx}$
& $\left ( \sum\limits_{x \in \prescript{l}{}{X}_i} f(x) \right ) ^ 2 $
& $ 1 $ \\

1st Order Taylor~\autocite{Molchanov}
& $A$
& $x\frac{d\loss}{dx}$
& $ \abs{ \sum\limits_{x \in \prescript{l}{}{X}_i} f(x) } $
& $n(\prescript{l}{}{X}_i)$ \\

1st Order Taylor, w. norm~\autocite{Molchanov}
& $A$
& $x\frac{d\loss}{dx}$
& $\abs{ \sum\limits_{x \in \prescript{l}{}{X}_i} f(x) } $
& $ \norm{\prescript{l}{}{\widetilde{S}} }_2 $ \\

Average of gradient~\autocite{Liu2019}
& $A$
& $\frac{d\loss}{dx}$
& $\sum\limits_{x \in \prescript{l}{}{X}_i} f(x) $
& $n(\prescript{l}{}{X}_i)$ \\

Collaborative channel pruning~\autocite{CollaborativeChannelPruning}
& $W$
& $\frac{1}{2} x_0 x_1 \frac{d^2\loss}{dx_0dx_1}$ (Table \ref{tab:approximations_taylor})
& $\sum\limits_{\substack{x_0, x_1 :\\ x_0, x_1 \notin \widetilde{\prescript{l}{}{X}_i}}} f(x_0, x_1)$
& $1$ \\

Connection sensitivity~\autocite{SNIP}
& $W$ (Section \ref{sec:connection_sensitivity})
& $x\diff{\loss}{x}$
& $\sum\limits_{x \in \prescript{l}{}{X}_i} \abs{f(x)}$
& $ \sum\limits_{s \in \prescript{l}{}{\widetilde{S}}} s $ \\

\midrule

\end{tabular}
\caption{Published approaches for channel pruning.}
\label{tab:classification_channel_pruning}
\end{center}
\end{table*}

\section{Weight based Saliency Metrics}

When computing weight-based metrics, all the information required to compute
the saliency is readily available.
The most common saliency metric used for pruning is the magnitude of the
weights. Specifically, the L2 and L1 norms of weights have been used in many
pruning schemes~\autocite{Dally, GroupWiseBrainDamage,Li2017,SoftFilterPruning,
StructuredProbabilisticPruning, LWC, DSD, AMC, Frankle_Carbin,
DynamicNetworkSurgery} for different granularities of pruning.  Magnitude-based
saliency metrics assume that weights of lower magnitude have a lesser
contribution to the network.

Another weight-based heuristic is \textit{min-weight}~\autocite{Molchanov}.
The sum of squared individual weights are scaled by the number of weights in
the channel to give the channel saliency (Equation \ref{eqn:min_weight}).

\begin{equation}
\prescript{l}{}{S}_i = \frac{1}{n(\prescript{l}{}{W}_i)} \sum\limits_{w \in \prescript{l}{}{W}_i} w ^2
\label{eqn:min_weight}
\end{equation}

Most weight-based metrics consider smaller weights to be less important to the
network.  For coarse granularities of pruning, one can also consider removing
redundant sets of weights. In the case of channel pruning, one can remove
channels that are similar to other channels.  Redundant channels can be removed
if they are not contributing to the final result.  He et
al.~\autocite{GeometricMedian} remove channels that have a euclidean distance
close to that of the layer's geometric median channel (see Equation
\ref{eqn:GeometricMedian}).

\begin{equation}
\prescript{l}{}{W}_{GM} \in argmin(g(x)) \text{ with }
g(x) = \sum\limits_{j=0}^{\prescript{l}{}{m}-1}\norm{x-\prescript{l}{}{W}_j}_2
\label{eqn:GeometricMedian}
\end{equation}

In this case the pointwise saliency $f(x)$ is given by $x - GM(x)$ where
$GM(x)$ is given according to Equation \ref{eqn:GeometricMedianPoint}.

\begin{equation}
\text{Given } x = \prescript{l}{}{W}_i[p, q, r] \text{, then } GM(x) = \prescript{l}{}{\mathcal{W}_{GM}}[p, q, r]
\label{eqn:GeometricMedianPoint}
\end{equation}

By removing channels that are close to the geometric median, one can assume that
they are already represented by the geometric median.  We denote
$reconstruction(x)$ any reconstruction of $x$ after pruning, then in this case
$reconstruction(x) = GM(x)$.

\subsubsection{Recursive Weight Based Metrics}
\label{sec:nisp}

Common weight-based methods only use weights from a single layer.  Saliency
metrics such as the L1-norm of weights treat channels as independent
components.

To illustrate this independence, let us consider the example of using the
L1-norm of weights as a saliency metric.  First, we compute the saliency metric
using the L1-norm of weights then remove the least salient channel.  We, then,
recompute the saliency of the remaining weights. The saliency of the channels
that were not pruned remain unchanged.

Now, let us consider the case where we use the L1-norm of output points.  We
compute the saliency of all the channels, prune the least salient channel and
finally recompute the saliency of the unpruned channels.  Let us assume that
the channel selected for pruning is from the third layer of the network.  The
saliency of the channels from the first, second and third layers (excluding the
pruned channel) are unchanged.  However, the saliency of channels from
following layers may have changed.  While the saliency of the channels are
computed in a independent way, the underlying information (the output points),
can be expressed recursively using the previous layers.  The recursive
component is implicit.  The recursive equation for outputs points is given in
Equation \ref{eqn:forward_pass} with $f^l$ being the forward pass function of
the $l^{th}$ layer.

\begin{equation}
\prescript{l}{}{A} = f^{l}(\prescript{l}{}{W}, \prescript{l-1}{}{A})
\label{eqn:forward_pass}
\end{equation}

Saliency metrics that use the gradients of the loss with respect to the weights
or the output points also have a component that is computed recursively using
backpropagation (see Equation \ref{eqn:backward_pass}.  Hence, the pointwise
metrics in Table \ref{tab:taxonomy} can use information from different layers.

\begin{equation}
\diff{\loss}{\prescript{l}{}{A}} = \diff{\loss}{\prescript{l+1}{}{A}} \diff{\prescript{l+1}{}{A}}{\prescript{l}{}{A}}
\label{eqn:backward_pass}
\end{equation}

Neuron Importance Score Propagation (NISP)~\autocite{NISP} uses an
explicitly recursive way of propagating saliency information between layers.

Equation \ref{eqn:nisp} shows the general propagation equation used in NISP
with $h^l$ being a function given by the authors.  $f^l$ depends only on the
type of the layer.

\begin{equation}
\prescript{l}{}{\mathcal{S}} = h^{l}(\prescript{l+1}{}{W}, \prescript{l+1}{}{\mathcal{S}})
\label{eqn:nisp}
\end{equation}

In this case the pointwise saliency $f(x)$ used by $NISP(x)$ is given in
Equation \ref{eqn:nispPoint}.

\begin{equation}
\text{Given } x = \prescript{l}{}{W}_i[p, q, r] \text{, then } NISP(x) = \prescript{l}{}{\mathcal{S}}[p, q, r]
\label{eqn:nispPoint}
\end{equation}

The method used by NISP can be considered as an alternative way of
backpropagating information through the network.

Gradient backpropagation and its alternatives can be used for pruning.  An
alternative to backpropagation of gradients is Layer-wise Relevance Propagation
(LRP)~\autocite{LRP}. LRP has successfully been used as a saliency metric for
pruning~\autocite{PruningByExplainingLRP}.  Similar to gradient backpropagation
and NISP, LRP propagates information recursively from the last (output) layer
of the network to its first (input) layer.  LRP requires weights and input
images as it is a data driven approach but can, nonetheless, be expressed in a
similar form to Equation \ref{eqn:nisp}.

We denote $alternateBackprop$ any alternative to backpropagation of information
in neural networks.  Hence, in the case of NISP we have $alternateBackprop(x) =
NISP(x)$.

\section{Weight and Input Images Based Saliency Metrics}

Channel pruning is a notable granularity of pruning as pruning an entire
channel of weights leads to the removal of a feature map from the network.
A given channel of weights that operates on a given input
produces an output feature map of outputs. It may be possible to identify good
candidates for pruning by selecting feature maps with low or zero outputs.

Output values across inferences on multiple inputs can be gathered, and
their results summarized using statistical measures. Saliency metrics use the
sum~\autocite{Anwar}, mean and variance~\autocite{Polyak_Wolf}, or
L2-norm~\autocite{Gaikwad} of feature maps to identify low saliency channels.

The feature map produced by the convolution layer is not the only feature map
that can be used. For example, the absolute percentage of zeros
(APoZ)~\autocite{Hu} counts the percentage of zero values in the output
activations for a given channel, and computes the average across multiple
inputs.

\begin{equation} \prescript{l}{}{S}_i =
 \frac{1}{n(\prescript{l}{}{A}_i)}  \sum _{a \in \prescript{l}{}{A}_i} a
\label{eqn:pruning_saliency_mean_act} \end{equation}

Quite often, convolution layers are followed by ReLU layers which only retain
positive outputs.  A negative mean would indicate that on average the outputs
produced by that channel were negative and likely to be driven to zeros by
ReLU.  Hence, APoZ considers the average of the output points \emph{after} ReLU.
APoZ considers channels with a higher percentage of zeroes to have a lower
saliency.

\begin{equation}
\prescript{l}{}{S}_i = \frac{1}{n( \prescript{l}{}{A}_i)}
\left ( \sum _{a \in \prescript{l}{}{A}_i} f(a) \right ) \text{with}
f(a) =
\begin{cases}
1& \text{if }a > 0\\
0 & \text{else }
\end{cases}
\label{eqn:pruning_saliency_apoz}
\end{equation}

A sub-category of saliency metrics that use weights and input images are
 metrics inspired from reconstruction error.  Saliency metrics that are
based on reconstruction error have a pointwise metric in the form of $x -
reconstruction(x)$. In the case of ThiNet~\autocite{ThiNet2019}, $x$ is a point sampled from the
input feature map and $reconstruction(x)$ is its reconstruction after pruning.
Metrics used by ThiNet~\autocite{ThiNet2019}, He et al.~\autocite{He2017}, and
Liu et al.  ~\autocite{DiscrimationAwarePruning} choose channels based on the
least error incurred to output feature maps.  Hence, the layer-wise feature
maps error is used as a saliency metric.  The main difference between the
approach explored by ThiNet~\autocite{ThiNet2019} and He et
al.~\autocite{He2017} is how they estimate the damaged feature map.  Liu et al.
~\autocite{DiscrimationAwarePruning} also introduce a layer-wise loss alongside
the reconstruction error.

\section{Weight, Input Images and Labels Based Saliency Metrics} \label{sec:gradient_based_methods}

Using only the weights and input images to compute saliency metrics does not
allow one to know directly how the classification performance is being
affected.  Corresponding labels are also needed to obtain this
information.  The \emph{loss} is a measure of how well the
predictions of the network match the ground truth.  Consequently, the gradients
with respect to the loss also carry this information.  A network that has
reached its minimum loss has a gradient of zero.

The equation of the cross-entropy loss, $\mathcal{L}$, is given by Equation
\ref{eqn:loss} for a network with weights $W$ and input dataset
$\mathcal{I}_{set}$. The dataset $\mathcal{I}_{set}$ contains $N$ pairs of
input images, $\prescript{}{n}{I}$, and labels (or true vector of probabilities
classifying $\prescript{}{n}{I}$), $\prescript{}{n}{L}$. $P$ gives the output
of the network, i.e., the vector of probabilities classifying the input image.

\begin{equation}
\loss(W, \mathcal{I}_{set}) =
\sum\limits_{n=0}^{N-1} \prescript{}{n}{\loss}(W) =
\sum\limits_{n=0}^{N-1} - (\prescript{}{n}{L} \odot log(P(W, \prescript{}{n}{I})))
\label{eqn:loss}
\end{equation}

$P$ is evaluated during a forward pass of the network using only the weights
and input images.  On the other hand evaluating $\loss$, requires the ground
truth and so do the gradients of the loss.  Hence, the use of the $loss$ and
its gradients carry more information than using only the weights or output
feature maps.  To compute the gradient of the loss, a backward pass as well as
a forward pass is required.

The use of gradients in saliency metrics for pruning was introduced by Mozer and
Smolensky's Skeletonization~\autocite{Mozer_Smolensky}, Lecun et al.'s Optimal
Brain Damage~\autocite{Lecun}, and Hassibi and Stork's Optimal Brain
Surgeon~\autocite{Hassibi}.

A simpler gradient based saliency measure was proposed by Liu and
Wu~\autocite{Liu2019}  where the average of the gradients of the output feature
maps (Equation \ref{eqn:saliency_average_activation_gradient}) is used as a
saliency measure of a channel.  They put forward that pruning channels that are
no longer updated by the SGD algorithm can be pruned safely.

\begin{equation} \prescript{l}{}{S}_i  = \frac{1}{n(\prescript{l}{}{A}_i)} \sum _{a \in
\prescript{l}{}{A}_i} \diff{\loss}{a}
\label{eqn:saliency_average_activation_gradient} \end{equation}

While Optimal Brain Damage introduced the use of Taylor expansions for deriving saliency
methods, other more recent approaches have also used Taylor expansions to
obtain different saliency metrics.  These methods are further explained in
section \ref{sec:taylor_expansion}

Saliency metrics that use a Taylor expansion estimate the error caused to the
final loss of the network.  Dong et al.~\autocite{Dong17} introduce a layer-wise
error, hence a layer-wise sensitivity and propose a method to propagate this
layer-wise sensitivity to deduce the final impact on the network. They use the
saliency measure introduced by Optimal Brain Surgeon~\autocite{Hassibi} to estimate the
layer-wise sensitivity.

\subsection{Connection sensitivity}
\label{sec:connection_sensitivity}

Lee et al.~\autocite{SNIP} define a saliency measure, Connection Sensitivity,  based on gradients of a mask term. Each
individual weight, $w_i$, have a mask term, $m_i$ , that can be either one or
zero.  Their saliency measure is derived using the gradients of the mask terms
instead of the gradients of the weights.

\begin{equation}
S(w_i) = \frac{\abs{ g_i (w; \mathcal{I}_{set}) }}{\sum\limits_{k=1}^{m}\abs{ g_k (w; \mathcal{I}_{set}) }}
\label{eqn:snip}
\end{equation}

\begin{equation}
g_i(w; \mathcal{I}_{set}) = \frac{\partial \loss (M \odot W; \mathcal{I}_{set})}{\partial m_i} \Bigr|_{M=1}
\label{eqn:snip_gradient}
\end{equation}

To facilitate the comparison of Connection Sensitivity to other saliency
metrics, we remind a few notations.  We use $W$ to denote the vector of the
weights and $M$, the vector of the mask terms. $M$ and $W$ are of similar
dimensions.  The vector of pruned weights, $\widetilde{W}$ with $i^{th}$ element
$\widetilde{w_i}$, is given by applying the mask on the weights with $\widetilde{W} = M
\odot W$ or $\widetilde{w_i} = m_i \cdot w_i$

Using this substitution, in Equation \ref{eqn:snip_gradient}, we obtain Equation
\ref{eqn:snip_gradient_sub}. The gradient of $\widetilde{W}$ with respect to the
loss, $\frac{d\loss}{d\widetilde{W}}$, is given by regular backpropagation rules.

In the case of Connection Sensitivity~\autocite{SNIP}, since $M=1$, i.e. all the components of M are set to 1, we
can express the gradients of the mask terms in terms of the known gradients of
the weights.  From Equation \ref{eqn:snip_gradient_eval}, we see that in this
case using the absolute value of the gradient of the mask, $\abs{g_i}$, is
similar to using the absolute value of the first term of a Taylor expansion,
$\abs{w_i \diff{\loss}{w_i}}$, from Equation
\ref{eqn:saliency_per_weight_diag_hessian}

\begin{eqnarray}
\label{eqn:snip_gradient_sub}
\frac{\partial \loss (M \odot W; \mathcal{I}_{set})}{\partial m_i} \Bigr|_{M=1}
& = \frac{\partial\loss (M \odot W; \mathcal{I}_{set})}{\partial \widetilde{w}_i} \cdot w_i \Bigr|_{M=1} \\
& = \frac{\partial \loss (W; \mathcal{I}_{set})}{\partial \widetilde{w}_i} \cdot \widetilde{w}_i \\
& = w_i \cdot \frac{\partial \loss (W; \mathcal{I}_{set})}{\partial w_i}
\label{eqn:snip_gradient_eval}
\end{eqnarray}

\subsection{Taylor Expansion} \label{sec:taylor_expansion}

Most of the gradient based saliency measures discussed in this paper can be
summarized as the estimation of the sensitivity of the parameters removed by
removing a convolutional filter channel from a network.  The sensitivity of a
parameter was first introduced for fully-connected layers
~\autocite{Mozer_Smolensky} as the change in the error of the network on the
training set caused by removing that parameter.  This definition can be
extended to convolution channels by using the change in error induced
by removing the set of parameters associated with that channel.  The
sensitivity of pruning a network with loss, $\mathcal{L}$, and unpruned
weights, $W$, to pruned weights, $\widetilde{W}$, is given in Equation
\ref{eqn:sensitivity}

\begin{equation}
Sensitivity = \loss(\widetilde{W}) - \loss(W)
\label{eqn:sensitivity}
\end{equation}

One of the first approaches to estimate the sensitivity of a weight was
proposed by Lecun et al.'s Optimal Brain Damage~\autocite{Lecun}. They use
 a simplified second order Taylor expansion on the trained neural network.  A
Taylor expansion is used to estimate the loss function, $\loss$, at the pruned
weights, $ \widetilde{W}$, using the trained weights, $W$.

\begin{eqnarray}
\nonumber
\loss(\widetilde{W})  \approx \loss(W)  &+& J (\widetilde{W} - W ) \\
&+& \frac{1}{2} (\widetilde{W} - W )^T H (\widetilde{W} - W )
\label{eqn:second_order_taylor_expansion}
\end{eqnarray}

A second order Taylor expansion around the trained weights, $W$, is given in
equation \ref{eqn:second_order_taylor_expansion}, where $J$ and $H$ are
respectively the Jacobian and Hessian of the loss function at trained
parameters $W$.  The Jacobian matrix, $J \in \mathcal{R}^{N_{weights}} $, is
defined as $J_i = \frac{d\loss}{dwi}$ and the Hessian matrix, $H \in
\mathcal{R}^{N_{weights} \times N_{weights}}$, is defined as $H_{ij} =
\frac{d^2\loss}{dw_idw_j}$

To prune an individual weight, ${w_i}$, from the network, ${w_i}$ is set
to zero, i.e. the $i^{th}$ parameter of $\widetilde{W}$ is set to zero.  To
easily understand how the saliency of a single parameter is derived, Equation
\ref{eqn:second_order_taylor_expansion} can be rewritten for pruning a single
parameter, $w_i$, in Equation
\ref{eqn:second_order_taylor_expansion_per_weight}.

\begin{eqnarray}
\nonumber
\loss(\widetilde{W})  &\approx& \loss(W)  + \frac{d\loss}{dw_i} (0 - w_i ) + \frac{1}{2} (0 - w_i ) \frac{d^2\loss}{d{w_i}^2} (0 - w_i ) \\
&\approx& \loss(W)   - w_i \frac{d\loss}{dw_i} + \frac{1}{2} {w_i}^2 \frac{d^2\loss}{d{w_i}^2}
\label{eqn:second_order_taylor_expansion_per_weight}
\end{eqnarray}

The approximation of the sensitivity given by the second order Taylor expansion
can be used as a saliency metric~\autocite{Lecun}.
Hence, the saliency of a single weight is given by Equation
\ref{eqn:saliency_per_weight_diag_hessian}.  Similarly, pruning a set of
parameters means setting these parameters to zero in  $\widetilde{W}$.

\begin{equation}
S(w_i)  =   - w_i\frac{\partial \loss}{\partial w_i}
            + \frac{1}{2}{w_i}^2\frac{{\partial}^2 \loss}{\partial {w_i}^2}
\label{eqn:saliency_per_weight_diag_hessian}
\end{equation}

Computing Equation \ref{eqn:second_order_taylor_expansion} exactly is very
expensive.  While the first order term of the equation (the term involving the
gradients) is computed in linear time, the higher order terms are more
difficult to obtain.  The Hessian matrix scales with the quadratic of the
number of weights in the network.  To better understand the difference in
computation and memory cost, let us consider the number of operations to
compute each element of a feature map through backpropagation.  During
backpropagation the gradients of the output feature map of a layer is given by
the next layer's backpropagation.  It is the gradients of the input feature map
that are computed during backpropagation. Given a layer $l$, the gradients
$\diff{\loss}{\prescript{l}{}{A}}$ are known, and
$\diff{\loss}{\prescript{l-1}{}A{}}$ is computed during the $l^{th}$ backward
pass.  Computing each point of $\diff{\loss}{\prescript{l-1}{}{A}}$ has a
complexity $O(\prescript{l}{}{m} \times (\prescript{l}{}{k})^2)$.  There are
$\prescript{l-1}{}{m} \times \prescript{l-1}{}{height} \times
\prescript{l-1}{}{width}$ such points to be computed for the gradient of the
input feature map.  To understand the higher complexity of computing the full
Hessian matrix, let us consider the computational cost of the \emph{layer-wise}
Hessian using chain rule.  If we did a full back propagation of the layer-wise
Hessian, we would instead need to compute $\frac{(\prescript{l-1}{}{m} \times
\prescript{l-1}{}{height} \times \prescript{l-1}{}{width})^2}{2}$ points each
having a complexity $O(\prescript{l}{}{m} \times (\prescript{l}{}{k})^4$).

\begin{figure}[]
\begin{center}
\includegraphics[width=1.0\columnwidth]{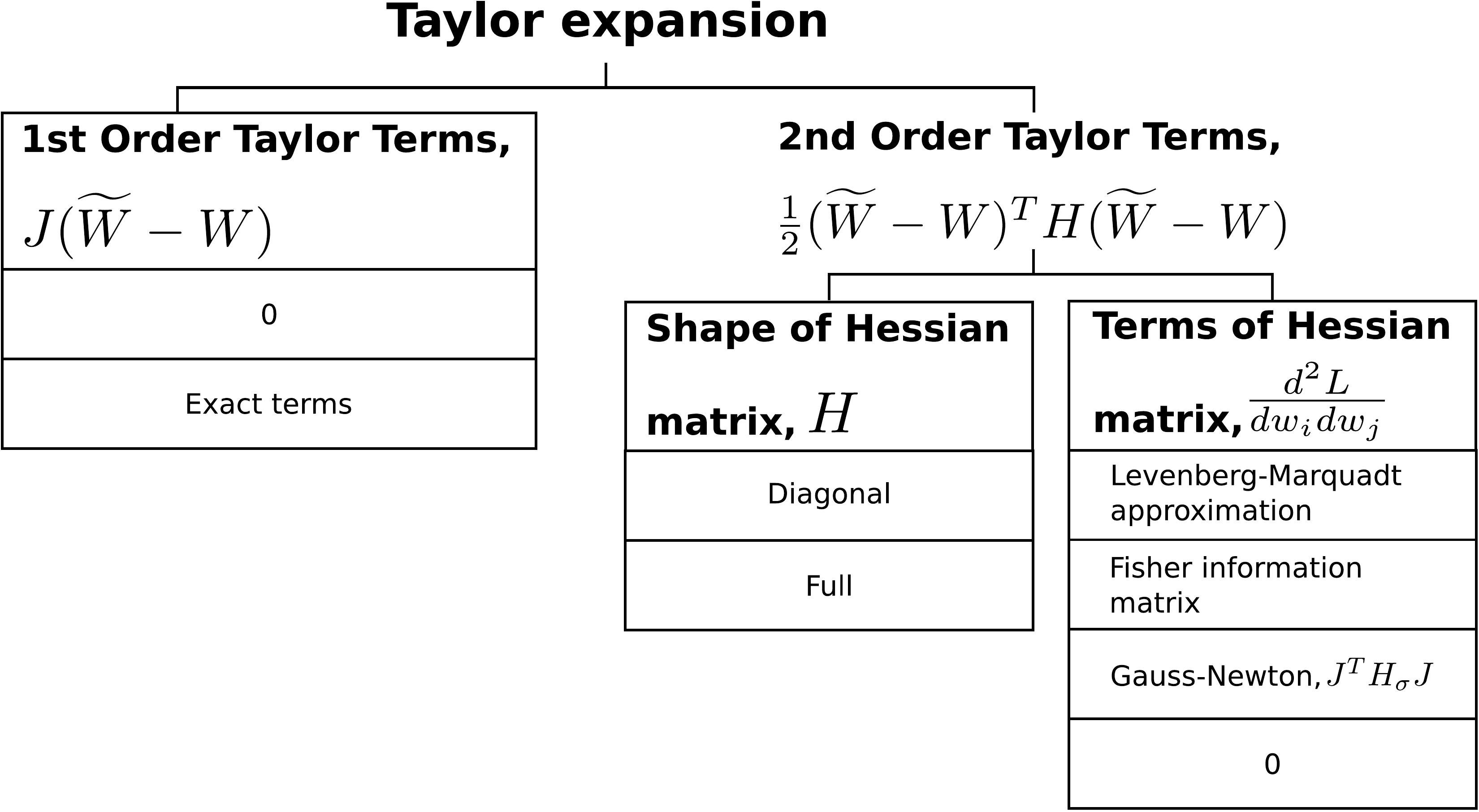}
\caption{Different approximations can be applied to Equation
\ref{eqn:second_order_taylor_expansion} to easily approximate the sensitivity of a set
of weights.}
\label{fig:approximations_taylor}
\end{center}
\end{figure}

To reduce
computation and storage cost, different approximations  can be applied to the
different terms of Equation \ref{eqn:second_order_taylor_expansion}
~\autocite{Lecun, Hassibi, Molchanov, Theis} to
obtain different saliency metrics. Popular approximations are presented in
Figure \ref{fig:approximations_taylor}.  Table \ref{tab:approximations_taylor}
summarizes various approximations of Equation
\ref{eqn:second_order_taylor_expansion} that have been used as saliency
metrics.

\begin{table*}[h]
\begin{center}
\scriptsize
\begin{tabular}{lc|rcr}
\multirow{2}{*}{\textbf{Saliency metric}} & \multirow{2}{*}{\textbf{1\s{st} order terms}} & \multicolumn{2}{c}{\textbf{2\s{nd} order terms (Hessian)}} \\
& & Shape & Approximate & Approximation Used \\
\midrule
Optimal Brain Damage~\autocite{Lecun} & Omitted & Diagonal & Y & Levenberg-Marquadt \\
Optimal Brain Surgeon~\autocite{Hassibi} & Omitted & Full & Y & Fisher\\
First order Taylor~\autocite{Molchanov} & Exact & Omitted & - & -\\
Fisher Information~\autocite{Theis} & Omitted & Diagonal & Y & Fisher\\
Collaborative Channel Pruning~\autocite{CollaborativeChannelPruning} & Omitted & Full & Y & Gauss-Newton with $H_{\sigma} = diag(L \oslash (P \odot P))$ \\
\end{tabular}
\caption{Approximations applied to the terms in Equation
\ref{eqn:second_order_taylor_expansion} to obtain a saliency metric for
pruning.}
\label{tab:approximations_taylor}
\end{center}
\end{table*}

Even though different approximations are used, the resulting saliency metrics
can be very similar. Equation \ref{eqn:fisher_theis} is used by Theis et
al.~\autocite{Theis} is very similar to Equation \ref{eqn:taylor_molchanov}
used by Molchanov et al.~\autocite{Molchanov}.

\begin{equation}
\prescript{l}{}{S}_i =
\frac{1}{n(\prescript{l}{}{A}_i)}
\abs{ \sum _{a \in \prescript{l}{}{A}_i}  a \diff{\loss}{a}}
\label{eqn:taylor_molchanov}
\end{equation}

\begin{equation}
\prescript{l}{}{S}_i = \frac{1}{2}
\left (
 \sum _{a \in \prescript{l}{}{A}_i}  a \diff{\loss}{a}
 \right ) ^2
\label{eqn:fisher_theis}
\end{equation}

%

\subsubsection{Consider only first order terms}
A first order expansion can also be used to approximate the change in loss.  The
second order terms in the Equation \ref{eqn:second_order_taylor_expansion} can
be set to zero to get the saliency metrics given by considering a first order
Taylor expansion.  The resulting equation has all its quantities readily
available during backpropagation.  The Jacobian matrix (i.e. the matrix of
gradients) is computed during backpropagation. 
Equation \ref{eqn:taylor_molchanov} is used by Molchanov et
al.~\autocite{Molchanov} as a saliency metric.

A first order Taylor approximation is theoretically a coarser
approximation of the real function than a second order Taylor expansion.
However, it has the advantage of omitting computation of higher order
derivatives.  In practice, we can see that good saliency metrics can be derived
from a first order Taylor expansion~\autocite{Molchanov,GlobalSparseMomentum}.

\subsubsection{Consider only second order terms}
On the other hand, other work choose to neglect the first order term.  A common
assumption is that if a network has been trained to a local minimum, its first
order derivatives will be very close to zero. This assumption about the
network's convergence, then allows the first order term (containing the
gradients) to be approximated to zero.  This a common approximation when using
second order Taylor expansions~\autocite{Lecun, Hassibi, Theis,
CollaborativeChannelPruning, Dong17}.

\subsubsection{Approximation of second order terms}
Computing the estimated terms using a second order Taylor expansion can be
expensive due to the cost of computing the Hessian matrix of the loss function
for every parameter.  To reduce computation cost of the Hessian matrix, the
terms that are not on its diagonal can be ignored~\autocite{Theis, Lecun}.
Considering only the diagonal terms of the Hessian reduces the number of points
to be computed.  The number of terms on the diagonal of the Hessian is equal to
the number of terms in the Jacobian (gradients).  To further reduce the
computation cost of the remaining terms, one can use more approximations. The
remaining terms Hessian can be estimated using a Levenberg-Marquardt
approximation for each layer of the network ~\autocite{Lecun}, the Fisher
information~\autocite{Theis} or a Gauss-Newton approximation.  The
Levenberg-Marquardt approximation used by Optimal Brain Damage~\autocite{Lecun}
propagate only the diagonal Hessian. Hence, its computational cost is similar
to backpropagating the gradients.  Optimal Brain Surgeon~\autocite{Hassibi,
Hassibi1993} also use the Fisher matrix as an approximation for the Hessian,
however they do not neglect the non-diagonal terms.  By including the
non-diagonal terms of the Hessian matrix, Optimal Brain
Surgeon~\autocite{Hassibi} consider the pairwise dependency between parameters.

\section{Experimental setup} \label{sec:setup}

\subsection{Saliency metrics}

We selected a subset of metrics that can be derived from Table
\ref{tab:taxonomy}, excluding $alternateBackprop$ (metrics based on an
alternative backpropagation) and $reconstruction$ (metrics based on
reconstruction error).

The second order Taylor expansion expressed in Table \ref{tab:taxonomy} cannot
be realistically computed exactly. The approximations seen in Figure
\ref{fig:approximations_taylor}  are used for the second order expansion.  We
choose to fix the shape of the Hessian to a diagonal matrix and use either an
expensive (Levenberg-Marquardt) or a cheap (Gauss-Newton with $H_{\sigma} = 1$)
algorithm to compute the remaining terms.  Neglecting the first order terms in
a second order Taylor expansion being a popular approximation~\autocite{Lecun,
Hassibi, Theis, CollaborativeChannelPruning}, we test this approximation,
leading to a total of 4 different pointwise saliency metrics that use the
Hessian.

We test all the saliency metrics that can be derived from the 2 base inputs, 7
different pointwise saliency metrics, 5 different reduction methods and 5
different scaling factors.  In practice, we obtain 308 saliency metrics with
different rankings for convolution channels.

\subsection{Pruning scheme}

Saliency metrics are embedded into a pruning scheme to remove weights from the
network.  Pruning schemes can range from simply removing a fixed number of
channels every iteration~\autocite{Dally}, to the use of reinforcement
learning~\autocite{AMC}, evolutionary particle filters~\autocite{Anwar} and
even genetic algorithms~\autocite{SparseLearningGeneticPruning}.  While
state-of-the-art pruning schemes push the boundaries of pruning further, simple
pruning schemes are still very efficient~\autocite{Dally,LWC,Frankle_Carbin}.
Simple pruning schemes heavily rely on how well the saliency metric can predict
the least salient entities.  Since our goal is to compare saliency metrics to
each other with the least number of confounding factors, we opt for simple
pruning schemes.
We evaluate the performance of the saliency metrics using the pruning
scheme given in Algorithm \ref{alg:experiment_retraining}.

We implement channel pruning so that at every step of the pruning process, we
have a dense network with fewer weights than before pruning. Using Algorithm
\ref{alg:experiment_retraining}, we select a channel to
prune from some convolutional layer of the network at each step. Where we would
remove a channel which participates in a join-type operation in directed
acyclic graph (DAG) structured networks (for example, pruning a channel from
one side of a skip connection in ResNet), we also remove the corresponding DAG
sibling channels, so that data dependencies are satisfied and the network
remains dense.  The weights of the dependent channels are included in
$\mathcal{TC}(\prescript{l}{}{W}_i)$.

\begin{algorithm}[h]
   \caption{Algorithm for pruning with retraining. Evaluating different channel
   selections for a CNN with loss function $\loss$, accuracy $\mathcal{Y}$ and
   converged weights $W$ with $M$ channels for a user-defined maximum drop in
   initial test accuracy, $testAccDrop$, maximum drop in train accuracy
   $trainAccDrop$ and maximum number of steps to use for retraining,
   $maxSteps$.}
   \label{alg:experiment_retraining}
\begin{algorithmic}
  \STATE $initialTestAcc \gets \mathcal{Y}(W, \mathcal{I}_{test})$
  \STATE $initialTrainAcc \gets \mathcal{Y}(W, \mathcal{I}_{train})$
  \REPEAT
    \STATE Compute $\prescript{l}{}{S}_i$, $ \forall l \in \{0 .. l_{max}-1 \}, \forall i \in \{ 0 .. \prescript{l}{}{m}-1 \}$ using $\mathcal{I}_{val}$

    \STATE Get $i$ and $l$, such that $\prescript{l}{}{S}_i = min(\prescript{p}{}{S}_q) $, $\forall p \in \{ 0 .. l_{max} - 1 \}, \forall q \in \{ 0 .. \prescript{l}{}{m}-1 \}$ and $\prescript{l}{}{W}_j$ is a non-zero tensor.
    \STATE $W \gets W - \mathcal{TC}(\prescript{l}{}{W}_j)$
    \STATE $retrainingSteps = 0$
    \REPEAT
      \STATE Retrain with 1 batch of images from $\mathcal{I}_{retrain}$
      \STATE $trainAcc \gets \mathcal{Y}(W, \mathcal{I}_{retrain})$
      \STATE $retrainingSteps ++$
    \UNTIL{($trainAcc > initialTrainAcc - trainAccDrop$) \\~~~~~~~or ($retrainingSteps \geq maxSteps$)}
    \STATE $testAcc \gets \mathcal{Y}(W, \mathcal{I}_{test})$
  \UNTIL{$testAcc < initialTestAcc - testAccDrop$}
\end{algorithmic}
\end{algorithm}

\subsection{Datasets}

We run our experiments using three different datasets: CIFAR-10
~\autocite{cifar10-paper}, CIFAR-100~\autocite{cifar10-paper} and a downsampled ImageNet
~\autocite{Chrabaszcz}.  For ImageNet-32, the images from ILSVRC 2012 challenge are
downsampled to $32 \times 32$ pixels~\autocite{Chrabaszcz}.  These three datasets all use $32
\times 32$ RGB input images with 10, 100 and 1000 different classes
respectively.

These three datasets each have their own disjoint training set, $\mathcal{I}_{train}$,
and testing set $\mathcal{I}_{test}$.  We train the CNNs on the whole training set,
$\mathcal{I}_{train}$, and measure their test accuracy using $\mathcal{I}_{test}$.

We split $\mathcal{I}_{train}$ into two disjoint sets $\mathcal{I}_{val}$ and $\mathcal{I}_{retrain}$.
$\mathcal{I}_{val}$ is used for computing the saliency metrics for channel pruning.
$\mathcal{I}_{retrain}$ is used only during the retraining phase.

\subsection{CNN models}

We conduct a wide range of experiments using different networks and different
datasets.  

We use the CIFAR-10~\autocite{cifar10-paper} dataset on LeNet, CIFAR10 network,
ResNet-20, NIN, and AlexNet.  We use ResNet-20~\autocite{resnet-paper}, and
NIN~\autocite{nin-paper} as originally described for the CIFAR-10
dataset.  We modify the first layer of LeNet-5~\autocite{lenet-paper} and
AlexNet ~\autocite{alexnet-paper} to process $32 \times 32$ RGB images instead
of their original input.

We use the CIFAR-100~\autocite{cifar10-paper} dataset on ResNet-20, NIN, and
AlexNet.  We use a downsampled ImageNet~\autocite{Chrabaszcz}, ImageNet-32, on
AlexNet.  The downsampled ImageNet contains the same number of images and
classes as original ImageNet~\autocite{imagenet-paper} but have each image resized to 32 by
32 pixels.

We maintain the  same input size for all our networks to 32 by 32 pixels.  If a
network is used for different datasets, the only structural change we apply to
that network is modifying the its last layer to classify either 10, 100 or 1000
classes. 

These nine networks are trained from scratch using Caffe~\autocite{caffe} and
their test accuracies are given in Table \ref{tab:sum_test_acc_cnn}.

\begin{table}[H]

\begin{center}
\scriptsize
\begin{tabular}{m{1.2cm}ccccc}
\multicolumn{1}{l}{}                 &                    \textbf{LeNet-5} & \textbf{CIFAR10} & \textbf{ResNet-20} & \textbf{NIN} & \textbf{AlexNet} \\
\multirow{2}{*}{\textbf{CIFAR-10}}   &  69.4\%           & 72.8\%           & 88.4\%             & 88.3\%       & 84.2\%           \\
\multirow{2}{*}{\textbf{CIFAR-100}}  &  -                & -                & 59.2\%             & 65.7\%       & 54.2\%           \\
\multirow{2}{*}{\textbf{ImageNet-32}}&  -                & -                & -                  & -            & 39.7\%           \\
\end{tabular}
\end{center}

\caption{Summary of trained network accuracy on CIFAR-10, CIFAR-100 and ImageNet-12.}
\label{tab:sum_test_acc_cnn}
\end{table}

\section{Results}

We begin by considering the three broad categories of saliency metrics proposed
in Figure~\ref{fig:information_taxonomy}. A naturally occuring question is to
identify the absolute best-performing method in each of our experimental
scenarios, and to determine the best pruning we can obtain of each network on
each dataset in experiments. 


In Table \ref{tab:average_best_weights}, we show the results for the best
performing saliency metric in each information category for each network. In
this evaluation, we compare the number of weights pruned allowing a drop in
TOP-1 accuracy of at most 5\%. When this threshold is exceeded, we stop the
experiment and take the last snapshot of the model which was above the
threshold as the candidate pruned model. We repeat this experiment for 8 runs,
and obtain the mean percentage of weights removed and a 95\% confidence
interval for the mean. This approach is used for all of the experimentation
presented.

\begin{table}[h]
\begin{center}
\caption{Effectiveness of metrics which use
different information.  The maximum sparsity achieved (\%)
with Algorithm \ref{alg:experiment_retraining} is shown.}
\label{tab:average_best_weights}
\scriptsize
\begin{tabular}{|l|c|c|c|}
\hline
\textbf{Network} & \textbf{Weights-based} & \textbf{Activation-based} & \textbf{Gradients-based}\\
\cline{2-4}
\hline
\multicolumn{4}{|l|}{\textbf{CIFAR-10 dataset}} \\
\hline
LeNet-5     &   80.4 $\pm$  2   & 78.5 $\pm$ 2  & \cellcolor{gray!25} \textbf{84.9} $\pm$ 4 \\
CIFAR-10    &   63.0 $\pm$  12  & 63.5 $\pm$ 4  & \cellcolor{gray!25} \textbf{67.5} $\pm$ 5 \\
ResNet-20   &   17.6 $\pm$  7   & 20.6 $\pm$ 24 & \cellcolor{gray!25} \textbf{25.4} $\pm$ 9 \\
NIN         &   63.6 $\pm$  2   & 59.1 $\pm$ 3  & \cellcolor{gray!25} \textbf{72.8} $\pm$ 3 \\
AlexNet     &   68.0 $\pm$  0.3 & 69.6 $\pm$ 2  & \cellcolor{gray!25} \textbf{70.1} $\pm$ 2 \\
\hline
\multicolumn{4}{|l|}{\textbf{CIFAR-100 dataset}} \\
\hline
ResNet-20   &   5.8 $\pm$  0.4                                &  6.6 $\pm$ 1   & \cellcolor{gray!25} \textbf{12.5} $\pm$ 3 \\
NIN         &   \cellcolor{gray!25} \textbf{59.2} $\pm$  1    & 54.4 $\pm$ 2   & 52.6 $\pm$ 1                              \\
AlexNet     &   60.2 $\pm$  0.1                               & 60.2 $\pm$ 11  & \cellcolor{gray!25} \textbf{64.0} $\pm$ 3 \\
\hline
\multicolumn{4}{|l|}{\textbf{ImageNet-32 dataset}} \\
\hline
AlexNet     &   \cellcolor{gray!25} \textbf{55.4} $\pm$ 7     & 51.6 $\pm$ 2   & 51.7 $\pm$ 5  \\
\hline
\end{tabular}
\end{center}
\end{table}

We can see that gradients-based methods are typically the best-performing in
experiments. However, there is no one standout method in our experiments, but
rather we see that different saliency metrics obtain the best results on
different networks or with different datasets. Table \ref{tab:average_best}
shows the saliency metrics corresponding to highlighted results in Table
\ref{tab:average_best_weights}.

Taking AlexNet as an example, we see that on each of the three classification
tasks, a different saliency metric produced the best results. On the CIFAR-10
and CIFAR-100 tasks, a gradient-based metric was ultimately most effective,
while on the ImageNet task, a pure weights-based metric won out.
The choice of \emph{which} gradients to consider also had an effect. For
CIFAR-10, a weight-oriented metric ($X = W$) was most effective, while on
CIFAR-100 an activation-oriented metric ($X=A$) was most effective.

\begin{table}[h]
\begin{center}
\scriptsize
\begin{tabular}{lC{1.5cm}l}
\textbf{Network} & \textbf{Sparsity \%}   & \textbf{Saliency Metric}   \\
\toprule
\multicolumn{3}{l}{\textbf{\scriptsize{CIFAR-10 dataset}}} \\
\midrule
\scriptsize{LeNet-5}
&  84.9 $\pm$ 4 &  $  \sum\limits_{x \in \prescript{l}{}{A}_i} \left ( \diff{\loss}{x} \right ) ^2 $\\
\scriptsize{CIFAR-10}
&  67.5 $\pm$ 5 &  $ \sum\limits_{x \in \prescript{l}{}{A}_i}\abs{-x\diff{\loss}{x} + \frac{x^2}{2}\ddiff{\loss}{x}_{GN}} $\\
\scriptsize{ResNet-20}
&  25.4 $\pm$ 9 &  $ \frac{1}{n(\prescript{l}{}{W}_i)}\left ( \sum\limits_{x \in \prescript{l}{}{A}_i} \diff{\loss}{x} \right ) ^2 $\\
\scriptsize{NIN}
&  72.8 $\pm$ 3 &  $ \frac{1}{n(\prescript{l}{}{W}_i)} \left ( \sum\limits_{x \in \prescript{l}{}{A}_i}   -x \diff{\loss}{x} + \frac{x^2}{2}\ddiff{\loss}{x}_{GN} \right )^2$\\
\scriptsize{AlexNet}
&  70.1 $\pm$ 2 &  $\frac{1}{n(\prescript{l}{}{W}_i)} \sum\limits_{x \in \prescript{l}{}{W}_i} \abs{-x\diff{\loss}{x}}$\\
\multicolumn{3}{l}{\textbf{\scriptsize{CIFAR-100 dataset}}} \\
\midrule
\scriptsize{ResNet-20}
&  12.5 $\pm$ 3 &  $ \left ( \sum\limits_{x \in \prescript{l}{}{A}_i} \diff{\loss}{x} \right ) ^2 $\\
\scriptsize{NIN}
&  59.2 $\pm$ 1 & $ \frac{1}{\norm{\prescript{l}{}{S}}_1}\sum\limits_{x \in \prescript{l}{}{W}_i} \abs{x}$ \\
\scriptsize{AlexNet}
&  64.0 $\pm$ 3 &  $\frac{1}{\norm{\prescript{l}{}{A}_i}_0} \sum\limits_{x \in \prescript{l}{}{A}_i} \abs{\diff{\loss}{x}}$\\
\multicolumn{3}{l}{\textbf{\scriptsize{ImageNet-32 dataset}}} \\
\midrule
\scriptsize{AlexNet}
&  55.4 $\pm$ 7 &  $\frac{1}{n(\prescript{l}{}{W}_i)} \sum\limits_{x \in \prescript{l}{}{W}_i} \abs{x}$\\
\end{tabular}
\caption{The best performing saliency metric in each scenario.}
\label{tab:average_best}
\end{center}
\end{table}
While these selected results show the best-performing metrics in our
experiments, we performed the same evaluation for all 308 candidate saliency
metrics. We cannot present data for all 308 experiments, but we summarize the
trends from the data in the remainder of this section as \emph{Findings}, which
highlight key trends with examples as appropriate. Only a small fraction of these
308 saliency metrics have been explored in previous literature. The data in
Tables~\ref{tab:average_best_weights} and ~\ref{tab:average_best} lead us to:



\begin{finding}
\label{finding:gradients_better}
\emph{Gradient-based metrics typically perform better than metrics which consider only weights or activations.}
\end{finding}

Metrics which use gradients require a full forward and backward pass of the
network to compute those gradients, as opposed to activation-based methods,
which require only a forward pass to compute activation values, or weight-based
metrics which require no computation beyond the application of the saliency
metric function to the weight values stored in the model. In our experiments,
gradient-based metrics yielded the best pruning or tied for the best pruning in
7 of 9 scenarios with different networks and datasets
(Table~\ref{tab:average_best_weights}).

\subsection{Weight-Based or Activation-Based Methods (Choice of $X$)}

In Table \ref{tab:pointwise_saliency_w} and \ref{tab:pointwise_saliency_a}, we
present the saliency achieved by each pointwise metric when $X=W$ and $X=A$
respectively.
From Table \ref{tab:pointwise_saliency_w} and \ref{tab:pointwise_saliency_a} we
see that when using pure weight-based metrics or activation-based metrics there
is not a clear cut winner. However, when the \emph{gradients} of either the
weights or output points are used, it is almost always preferable to use the
gradients of output points.

\begin{finding}
\label{finding:activation_better}
\emph{Pruning using the gradient with respect to the output points often outperforms pruning using the gradient with respect to the weights.}
\end{finding}

With channel pruning, there is a one-to-one correspondence between groups of
weights and groups of output points. However, at finer granularities there is no
one-to-one correspondence, but rather a one-to-many correspondence. With this
mixing of information at finer granularities, we expect the gradient with
respect to output points to be a less reliable signal for non-channel-oriented
pruning.



\subsection{Pointwise metric (choice of $F$)}



In order to compare pointwise saliency metrics, we need to fix the reduction
$R$ and scaling $K$ to reasonable choices which can be expected to give good
results on average. Equation \ref{eqn:comp_reduction_scaling} shows a common
choice in the literature~\autocite{Dally, Li2017, StructuredProbabilisticPruning,
LWC, AMC, DynamicNetworkSurgery, GlobalSparseMomentum}. Here the reduction $R$
is the L1-norm, and the scaling factor $K = 1$.


\begin{equation}
\prescript{l}{}{S}_i = \sum\limits_{x \in \prescript{l}{}{X}_i} \abs{f(x)}
\label{eqn:comp_reduction_scaling}
\end{equation}

For constructing pointwise metrics, the use of a Taylor expansion around the
loss function is very common in the literature. We tested five Taylor
expansions which use different approximations. From the results in Tables
\ref{tab:pointwise_saliency_a} and \ref{tab:pointwise_saliency_w}, we observe
that some approximations are often poor. In particular, neglecting the first
order terms (the gradient) when using a second order Taylor expansion around
the loss function is a poor approximation in most cases. The
degree to which the training process has \emph{converged} before pruning
affects the magnitudes of gradients, meaning they may still be quite large in
many cases, so assuming that they are universally close to zero can be a very
coarse approximation.

\begin{finding}
\emph{First order terms are often not negligible in second order Taylor expansions.}
\end{finding}


The second order term in the Taylor expansion $ -x\diff{\loss}{x} +
\frac{x^2}{2}\ddiff{\loss}{x} $ corresponds to the Hessian of the loss
function, which is very expensive to compute. Several approximations of the
Hessian have been used in the literature. Using a Levenberg-Marquardt
approximation requires a very expensive backward propagation of the second
order derivatives. This is not commonly implemented in popular deep learning
frameworks, because only the first derivative (i.e. the gradient) is required
for training. We found that, in the majority of cases in our experiments, a
Gauss-Newton appproximation of the Hessian is sufficient for pruning. In
Tables~\ref{tab:pointwise_saliency_w} and ~\ref{tab:pointwise_saliency_a}, we see
that the Levenberg-Marquadt approximation is only clearly advantageous in one
case, when pruning NIN on CIFAR-100.

\begin{finding}
\emph{The Gauss-Newton approximation of the Hessian is sufficiently accurate for pruning.}
\end{finding}

\begin{table*}[]
\begin{center}
\caption{Comparison between different pointwise saliency metrics. The maximum
proportion of weights removed (sparsity) by Algorithm \ref{alg:experiment_retraining} (\%) using different pointwise saliency metrics with
weights as the input ($x=w$) and Equation \ref{eqn:comp_reduction_scaling} as
reduction method.}
\label{tab:pointwise_saliency_w}
\scriptsize
\begin{tabular}{|l|l|l|l|l|l|c|l|}
\hline
\textbf{Metric}
& $w$
& $\diff{\loss}{w}$
& $-w\diff{\loss}{w}$
& $-w\diff{\loss}{w} + \frac{w^2}{2}\ddiff{\loss}{w}_{GN}$
& $\frac{w^2}{2}\ddiff{\loss}{w}_{GN}$
& $-w\diff{\loss}{w} + \frac{w^2}{2}\ddiff{\loss}{w}_{LM}$
& $\frac{w^2}{2}\ddiff{\loss}{w}_{LM}$
\\
\hline
\multirow{4}{*}{\textbf{Network}}

 & \multirow{4}{1.5cm}{\textbf{Weights only}} & \multicolumn{6}{l|}{\textbf{Weights, input images and labels}}  \\
 \cline{3-8}
 & & \multirow{3}{1.5cm}{\textbf{Gradients only}} & \multicolumn{5}{l|}{\textbf{Taylor expansions}} \\
 \cline{4-8}
 & & & \multirow{2}{*}{\textbf{1\s{st} order}} & \multicolumn{4}{l|}{\textbf{2\s{nd} order with diagonal Hessian}}  \\
 \cline{5-8}
 & & & & \multicolumn{2}{l|}{\textbf{Gauss-Newton approximation}} & \multicolumn{2}{l|}{\textbf{Levenberg-Marquardt approximation}} \\
\cline{1-8}
\multicolumn{8}{|l|}{\textbf{ CIFAR-10 dataset}} \\
\hline
LeNet-5 & 78.7 $\pm$ 9.8 & 64.3 $\pm$ 13.9 & \cellcolor{gray!25} \textbf{80.6} $\pm$ 3.0 & 80.3 $\pm$ 3.8 & 79.0 $\pm$ 8.0 & \cellcolor{gray!25} \textbf{80.8} $\pm$ 2.8 & 78.5 $\pm$ 5.2\\
CIFAR10 & 16.7 $\pm$ 62.9 & 19.3 $\pm$ 19.7 & 53.0 $\pm$ 24.5 & 50.7 $\pm$ 24.2 & \cellcolor{gray!25} \textbf{61.7} $\pm$ 17.6 & 50.4 $\pm$ 28.7 & 22.3 $\pm$ 5.6\\
ResNet-20 & 1.7 $\pm$ 0.0 & \cellcolor{gray!25} \textbf{6.7} $\pm$ 8.7 & 4.6 $\pm$ 3.7 & 4.6 $\pm$ 2.8 & 4.0 $\pm$ 2.3 & 4.4 $\pm$ 4.2 & 1.2 $\pm$ 0.1\\
NIN & 34.5 $\pm$ 0.5 & 5.6 $\pm$ 1.8 & \cellcolor{gray!25} \textbf{61.1} $\pm$ 2.8 & 20.1 $\pm$ 59.1 & 13.7 $\pm$ 71.5 & 60.3 $\pm$ 2.3 & 42.8 $\pm$ 1.7\\
AlexNet & \cellcolor{gray!25} \textbf{64.0} $\pm$ 9.6 & 40.1 $\pm$ 4.8 & 51.7 $\pm$ 9.9 & 52.0 $\pm$ 9.8 & 49.7 $\pm$ 8.6 & 55.1 $\pm$ 6.1 & 20.0 $\pm$ 24.8\\

\hline
\multicolumn{8}{|l|}{\textbf{ CIFAR-100 dataset}} \\
\hline
ResNet-20 & 3.4 $\pm$ 0.2 & 3.2 $\pm$ 3.8 & 2.8 $\pm$ 6.1 & \cellcolor{gray!25} \textbf{4.1} $\pm$ 5.0 & 1.9 $\pm$ 14.0 & \cellcolor{gray!25} \textbf{4.1} $\pm$ 3.4 & 1.1 $\pm$ 0.1\\
NIN & 42.2 $\pm$ 1.0 & 35.2 $\pm$ 0.2 & 36.2 $\pm$ 0.1 & 36.4 $\pm$ 1.0 & 37.1 $\pm$ 0.1 & 36.2 $\pm$ 0.9 & \cellcolor{gray!25} \textbf{52.4} $\pm$ 2.7\\
AlexNet & \cellcolor{gray!25} \textbf{58.6} $\pm$ 21.4 & 26.4 $\pm$ 14.5 & 56.1 $\pm$ 14.8 & 52.0 $\pm$ 19.3 & 50.2 $\pm$ 11.2 & 52.1 $\pm$ 5.7 & 27.0 $\pm$ 7.3\\

\hline
\multicolumn{8}{|l|}{\textbf{ ImageNet-32 dataset}} \\
\hline
AlexNet & 28.2 $\pm$ 2.1 & 31.0 $\pm$ 3.3 & 31.1 $\pm$ 0.8 & 31.2 $\pm$ 2.2 & \cellcolor{gray!25} \textbf{35.2} $\pm$ 1.0 & 31.2 $\pm$ 0.8 & 1.1 $\pm$ 0.1\\
\hline
\end{tabular}
\end{center}
\end{table*}

\begin{table*}[]
\begin{center}
\caption{Comparison between different pointwise saliency metrics. The maximum
proportion of weights removed (sparsity) by Algorithm \ref{alg:experiment_retraining} (\%) using different pointwise saliency metrics with
output points as the input ($x=a$) and Equation \ref{eqn:comp_reduction_scaling} as
reduction method.}
\label{tab:pointwise_saliency_a}
\scriptsize
\begin{tabular}{|l|l|l|l|l|l|c|l|}

\hline

\textbf{Metric}
& $a$
& $\diff{\loss}{a}$
& $-a\diff{\loss}{a}$
& $-a\diff{\loss}{a} + \frac{a^2}{2}\ddiff{\loss}{a}_{GN}$
& $\frac{a^2}{2}\ddiff{\loss}{a}_{GN}$
& $-a\diff{\loss}{a} + \frac{a^2}{2}\ddiff{\loss}{a}_{LM}$
& $\frac{a^2}{2}\ddiff{\loss}{a}_{LM}$ \\
\hline

\multirow{4}{*}{\textbf{Network}}

 & \multirow{4}{1.5cm}{\textbf{Weights and input images}} & \multicolumn{6}{l|}{\textbf{Weights, input images and labels}}  \\
 \cline{3-8}
 & & \multirow{3}{1.5cm}{\textbf{Gradients only}} & \multicolumn{5}{l|}{\textbf{Taylor expansions}} \\
 \cline{4-8}
 & & & \multirow{2}{*}{\textbf{1\s{st} order}} & \multicolumn{4}{l|}{\textbf{2\s{nd} order with diagonal Hessian}}  \\
 \cline{5-8}
 & & & & \multicolumn{2}{l|}{\textbf{Gauss-Newton approximation}} & \multicolumn{2}{l|}{\textbf{Levenberg-Marquardt approximation}} \\
\cline{2-8}
\hline
\multicolumn{8}{|l|}{\textbf{ CIFAR-10 dataset}} \\
\hline
LeNet-5 & 77.1 $\pm$ 2.5 & 82.0 $\pm$ 2.1 & 82.5 $\pm$ 2.2 & \cellcolor{gray!25} \textbf{82.6} $\pm$ 2.7 & 74.6 $\pm$ 2.3 & 82.1 $\pm$ 1.8 & 69.7 $\pm$ 5.6\\
CIFAR10 & 56.0 $\pm$ 14.5 & 56.7 $\pm$ 19.6 & 65.4 $\pm$ 15.0 & \cellcolor{gray!25} \textbf{66.8} $\pm$ 16.0 & 57.3 $\pm$ 21.6 & 65.3 $\pm$ 15.9 & 22.3 $\pm$ 17.2\\
ResNet-20 & 5.5 $\pm$ 3.2 & 4.2 $\pm$ 3.5 & 11.5 $\pm$ 14.6 & 12.8 $\pm$ 17.3 & 9.5 $\pm$ 14.1 & \cellcolor{gray!25} \textbf{13.2} $\pm$ 13.4 & 2.4 $\pm$ 0.3\\
NIN & 32.9 $\pm$ 25.4 & 5.4 $\pm$ 61.8 & 58.3 $\pm$ 4.3 & 56.9 $\pm$ 4.1 & \cellcolor{gray!25} \textbf{62.7} $\pm$ 14.2 & 59.7 $\pm$ 12.6 & 38.7 $\pm$ 24.6\\
AlexNet & \cellcolor{gray!25} \textbf{69.2} $\pm$ 4.4 & 63.9 $\pm$ 5.6 & 65.0 $\pm$ 3.0 & 63.3 $\pm$ 2.4 & 57.0 $\pm$ 5.4 & 63.8 $\pm$ 4.3 & 35.3 $\pm$ 5.1\\

\hline
\multicolumn{8}{|l|}{\textbf{ CIFAR-100 dataset}} \\
\hline
ResNet-20 & 3.8 $\pm$ 1.7 & 2.4 $\pm$ 1.0 & 4.6 $\pm$ 2.4 & 4.7 $\pm$ 2.8 & \cellcolor{gray!25} \textbf{5.2} $\pm$ 2.3 & 4.6 $\pm$ 1.7 & 6.4 $\pm$ 4.3\\
NIN & 42.6 $\pm$ 2.2 & 35.2 $\pm$ 0.0 & 36.2 $\pm$ 0.1 & 36.2 $\pm$ 0.0 & 37.0 $\pm$ 0.8 & 36.1 $\pm$ 0.1 & \cellcolor{gray!25} \textbf{45.6} $\pm$ 2.0\\
AlexNet & 57.6 $\pm$ 6.7 & \cellcolor{gray!25} \textbf{62.9} $\pm$ 2.2 & 62.7 $\pm$ 3.3 & \cellcolor{gray!25} \textbf{62.9} $\pm$ 2.7 & 52.5 $\pm$ 5.8 & 62.4 $\pm$ 3.7 & 33.1 $\pm$ 23.6\\

\hline
\multicolumn{8}{|l|}{\textbf{ ImageNet-32 dataset}} \\
\hline
AlexNet & 19.4 $\pm$ 4.1 & \cellcolor{gray!25} \textbf{40.4} $\pm$ 1.9 & 30.3 $\pm$ 1.2 & 30.2 $\pm$ 1.5 & 25.6 $\pm$ 0.9 & 30.4 $\pm$ 1.7 & 2.9 $\pm$ 1.4\\
\hline

\end{tabular}
\end{center}
\end{table*}

\subsection{Reduction (choice of $R$)}

When pruning blocks of weights, the pointwise metrics are combined into a
single value using some reduction function. Existing research on pruning
often places little focus on reduction and scaling methods, and it can
sometimes be difficult to identify the approach used in any given method.
Nonetheless, the method by which the pointwise metric is reduced and scaled can greatly
influence the quality of a saliency metric.


\begin{figure*}[h]
\centering

\subfloat[][For all tested $X$ and $f(x)$.]{
\label{fig:reduction_scaling_summary}
\includegraphics[width=0.5\textwidth]{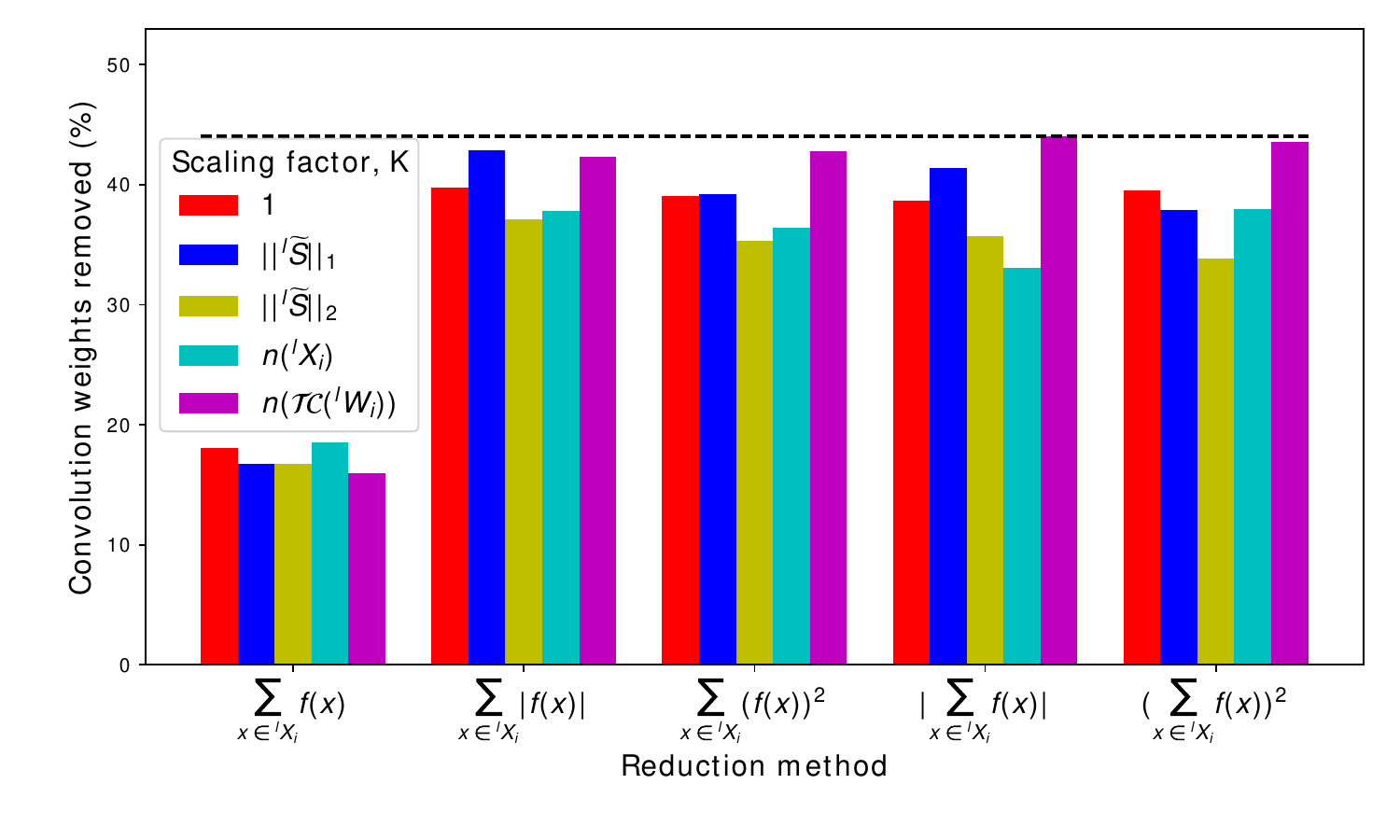} }
\subfloat[][$X=W$ and $f(x) = x$]{
\label{fig:reduction_scaling_w_avg}
\includegraphics[width=0.5\textwidth]{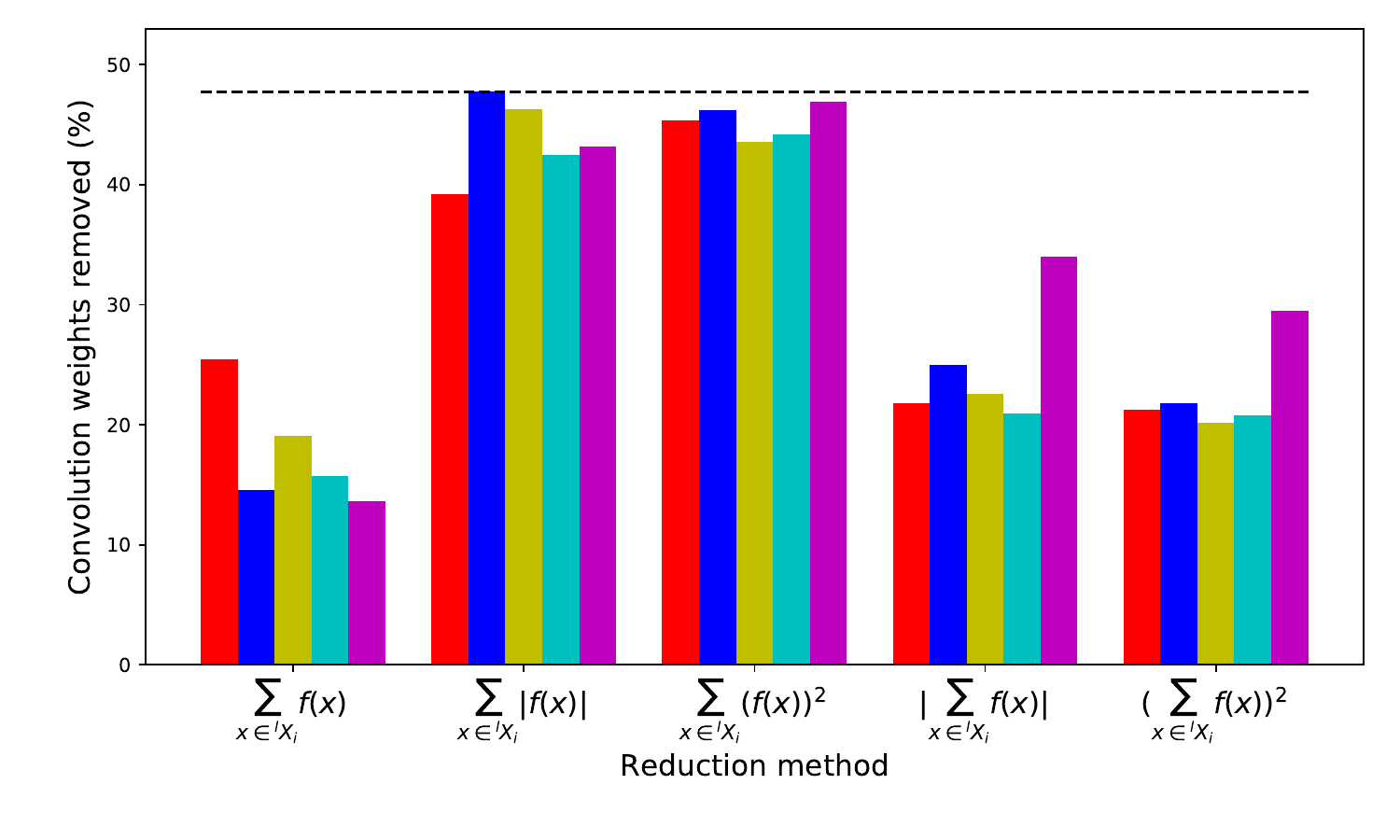} }
\\
\subfloat[][$X=A$ and $f(x) = x$]{
\label{fig:reduction_scaling_a_avg}
\includegraphics[width=0.5\textwidth]{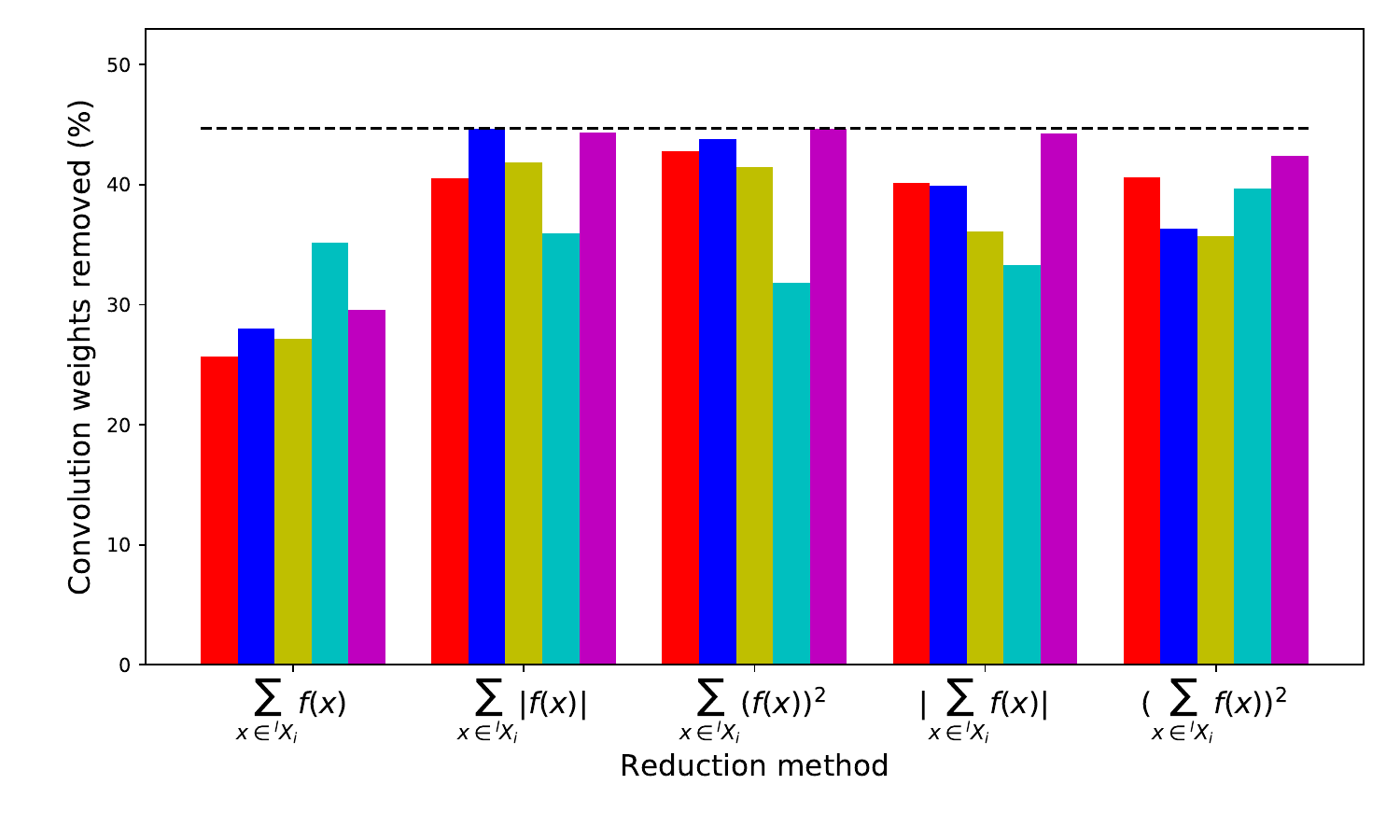} }
\subfloat[][$X=A$ and $f(x) = -x\diff{\loss}{x} + \frac{x^2}{2}\ddiff{\loss}{x}_{GN}$]{
\label{fig:reduction_scaling_a_t22}
\includegraphics[width=0.5\textwidth]{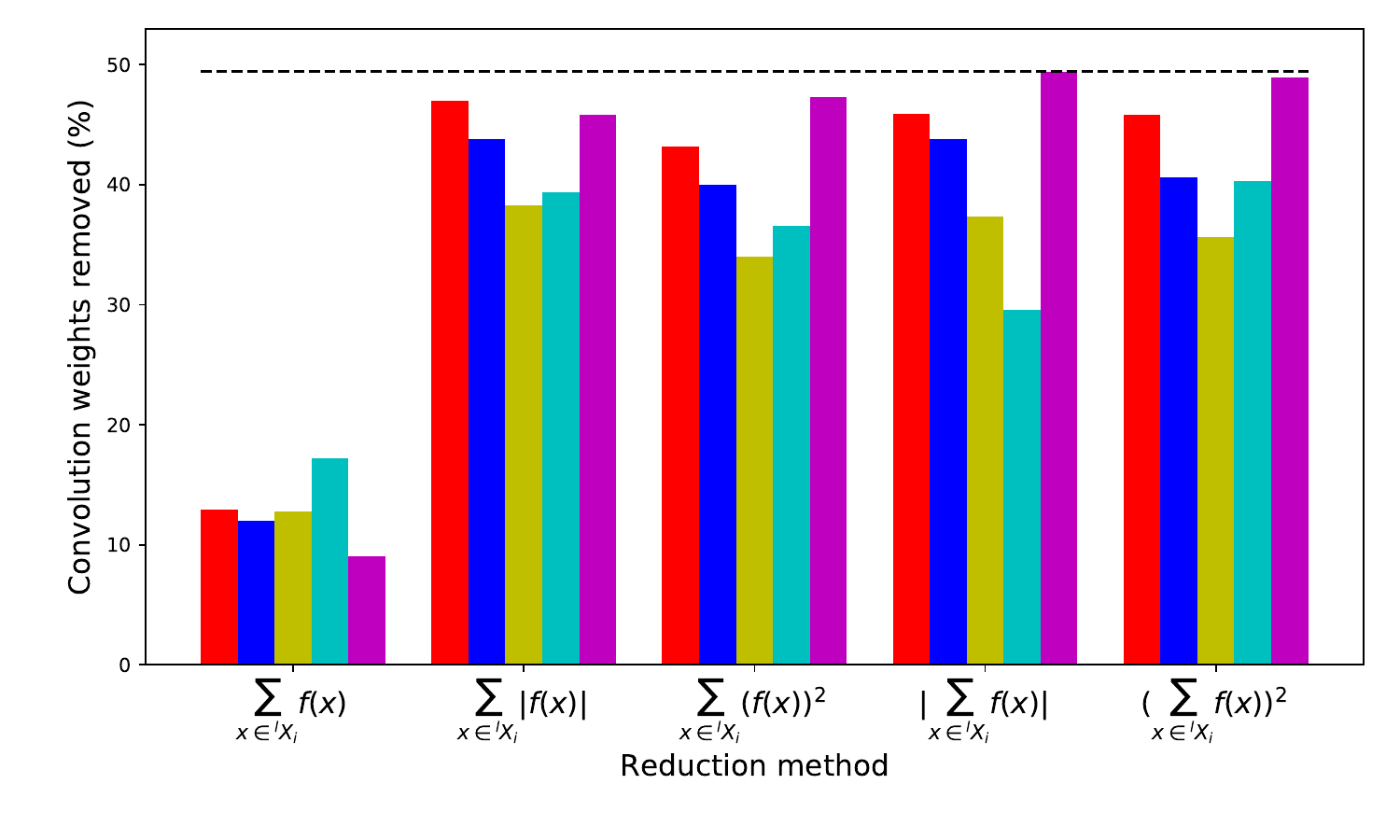} }

\caption{Comparison between different reduction methods and scaling factors.
The percentage of convolution weights removed (sparsity) on average for different scaling
methods across different networks and datasets.}
\label{fig:reduction_scaling}.
\end{figure*}

\begin{finding}
\label{finding:positive_metrics}
\emph{Strictly positive saliency metrics offer better pruning results.}
\end{finding}

Figure \ref{fig:reduction_scaling} presents the results of our experimentation
grouped so that only the reduction and scaling methods vary.
Figure~\ref{fig:reduction_scaling_summary} presents a summary for all choices
of input and all pointwise metrics. We then examine the three families of
metrics outlined in Figure~\ref{fig:information_taxonomy} in more detail:
metrics which consider static information (i.e. weights) only ($X=W$,
Figure~\ref{fig:reduction_scaling_w_avg}), metrics which consider information
available with only a forward pass, i.e. weights and output feature maps,
($X=A$, Figure~\ref{fig:reduction_scaling_a_avg}) and finally metrics which
consider all information available from a forward and backward pass ($X=A$
and $f(x)$ involves a gradient, Figure~\ref{fig:reduction_scaling_a_t22}).

As a general trend, we see that using the same pointwise saliency metric
but varying the reduction or scaling methods can produce very different
results. For example, we see that using the raw sum of the gradients (first bar
set in each graph) typically results in poorly performing metrics versus other
reduction methods. In each scenario, reducing by the sum of the \emph{absolute}
values of the gradients produces significantly better results (second bar set
in each graph).

We observe that the gap between the simple summation and the other
guaranteed-positive reduction methods is smaller in Figure
\ref{fig:reduction_scaling_a_avg} where we use $X = A$, i.e. the gradient with
respect to output features. In our experiments, networks containing ReLU layers
are optimized such that ReLU is \emph{fused} with the convolution layer,
meaning the resulting output features are non-negative. This means the simple
summation is also guaranteed non-negative, which improves the quality of
pruning decisions significantly.

%

\subsection{Scaling (choice of $K$)}

In Figures \ref{fig:reduction_scaling_w_avg} and
\ref{fig:reduction_scaling_a_avg} we see that two scaling methods stand out
when we ignore the typically poorly-performing first bar group, which
corresponds to the simple-summation reduction method. Across all four remaining
guaranteed-positive reduction methods, two scaling factors trade blows for
first and second place in terms of the quality of pruning decisions:
$\norm{\prescript{l}{}{\widetilde{S}}}_1$ and
$n(\mathcal{TC}(\prescript{l}{}{W}_i))$.

Recall from Section~\ref{sec:scaling} that
$\norm{\prescript{l}{}{\widetilde{S}}}_1$ is the layer-wise L1 norm of
saliency values, and $\mathcal{TC}(\prescript{l}{}{W}_i)$ denotes the entire
set of weights \emph{transitively} removed when $\prescript{l}{}{W}_i$ is
removed from the network. Both of these scaling factors incorporate structural
information, as opposed to strictly local information about the parameters
being pruned.

\begin{finding}
\label{finding:transitive}
\emph{Incorporating structural information of the network in the scaling factor
offers better pruning results.}
\end{finding}

In Figure \ref{fig:reduction_scaling_w_avg}, we can directly compare scaling by
the local ($n(\prescript{l}{}{W}_i)$) and transitive
($n(\mathcal{TC}(\prescript{l}{}{W}_i))$) number of weights removed. These are
the second-to-last and last bars in each bar set, respectively. While the
improvement of using $n(\mathcal{TC}(\prescript{l}{}{W}_i))$ is sometimes
small, it is strictly better than using only the local information in each
case when considering guaranteed-positive reduction methods.

In Figure \ref{fig:reduction_scaling_a_t22}, we look at how a good
gradient-based pointwise metric can be affected by reduction and scaling.
Similarly to non-gradient-based metrics in
Figures~\ref{fig:reduction_scaling_w_avg}
and~\ref{fig:reduction_scaling_a_avg}, we see that the use of a reduction
method that is guaranteed positive led to significantly better results.  The
gradient-based metric results further highlight the benefit of using non-local
information. The best overall pruning result is achieved using
$n(\mathcal{TC}(\prescript{l}{}{W}_i))$ as the scaling factor (fourth bar set
in the figure), but using the local scaling factor $n(\prescript{l}{}{W}_i)$
while keeping everything else fixed makes the quality of the metric plummet --
in this case, by nearly 20 percentage points average sparsity achieved across
all networks and datasets in our experiments.

\begin{finding}
\label{finding:transitive_weights}
\emph{The number of weights transitively removed is a better scaling
factor better than the number of weights locally removed.}
\end{finding}




\subsection{Saliency Metrics and Retraining Iterations}
\label{sec:retraining}

Retraining or fine-tuning is a crucial step in many pruning
algorithms because it allows the network to adjust remaining parameters
to compensate for the damage done by the removal of pruned
parameters. However, when evaluating saliency metrics, retraining
is a confounding factor because it can arbitrarily change weight values.
In fact, the effect of retraining is so great that we can often compensate for
the suboptimal choices made by poor saliency metrics with enough retraining.

Confounding factors notwithstanding, it seems intuitive that a better
saliency metric, should greatly reduce the effort spent on retraining
to achieve a given target accuracy and sparsity.
Better saliency metrics do less damage to the network to attain
a given threshold minimum sparsity, and conversely also result in higher
achievable sparsity ratios for a given threshold minimum accuracy. Thus, we
expect the total computational cost of pruning (retraining included) to be reduced by
choosing a higher-quality saliency metric.

\begin{table*}[]
\begin{center}
\caption{Spearman correlation (with p-value) between metric quality and computational cost of pruning including retraining.}
\label{tab:correlation}
\scriptsize
\begin{tabular}{|l|l|l|l|l|l|l|l|l|}
\hline
\multicolumn{5}{|l|}{\textbf{\scriptsize{CIFAR-10}}}
&\multicolumn{3}{l|}{\textbf{\scriptsize{CIFAR-100}}}
&\textbf{\scriptsize{ImageNet-32}} \\
\hline
\textbf{LeNet-5} & \textbf{CIFAR10} & \textbf{ResNet-20} & \textbf{NIN} & \textbf{AlexNet}  & \textbf{ResNet-20} & \textbf{NIN} & \textbf{AlexNet}& \textbf{AlexNet}\\
\hline
-0.10
($3e^{-1}$)
& -0.5
($8e^{-4}$)
& 1
($0$)
& -0.7
($6e^{-2}$)
& -0.9
($2e^{-19}$)
& -0.6
($1e^{-1}$)
& -0.6
($4e^{-1}$)
& -0.8
($4e^{-18}$)
& -0.9
($3e^{-3}$) \\
\hline
\end{tabular}
\end{center}
\end{table*}

\begin{algorithm}[h]
   \caption{Algorithm for pruning without retraining. Evaluating different
   channel selections for a CNN with loss function $\loss$, accuracy
   $\mathcal{Y}$ and converged weights $W$ with $M$ channels for a user-defined
   maximum drop in initial test accuracy, $maxTestAccDrop$.}
   \label{alg:experiment}
\begin{algorithmic}
  \STATE $initialTestAcc \gets \mathcal{Y}(W, \mathcal{I}_{test})$
  \REPEAT
    \STATE Compute $\prescript{l}{}{S}_i$, $ \forall l \in \{0 .. l_{max}-1 \}, \forall i \in \{ 0 .. \prescript{l}{}{m}-1 \}$ using $\mathcal{I}_{val}$
    \STATE Get $i$ and $l$, such that $\prescript{l}{}{S}_i = min(\prescript{p}{}{S}_q) $, $\forall p \in \{ 0 .. l_{max} - 1 \}, \forall q \in \{ 0 .. \prescript{l}{}{m}-1 \}$ and $\prescript{l}{}{W}_j$ is a non-zero tensor.
    \STATE $W \gets W - \mathcal{TC}(\prescript{l}{}{W}_j)$
    \STATE $testAcc \gets \mathcal{Y}(W, \mathcal{I}_{test})$
  \UNTIL{$testAcc < initialTestAcc - testAccDrop$}
\end{algorithmic}
\end{algorithm}

We examine the relationship between the quality of a saliency metric and the
total computational cost of pruning. As a proxy for the quality of a saliency metric we use
the maximum sparsity achieved using that metric \emph{without} any retraining,
i.e. using Algorithm \ref{alg:experiment}.  The pruning scheme outlined in
Algorithm \ref{alg:experiment}, is similar to our previous pruning scheme,
except for the omission of retraining steps.

We refer to the total cost of pruning as the total computational cost to reach a certain sparsity
while maintaining a fixed accuracy target.  Hence, in addition to the accuracy
threshold in Algorithm \ref{alg:experiment_retraining}, we add a stopping
condition related to the sparsity.  When this target is met we stop the
experiment and record the total number of steps that were required to prune and
retrain the network. The target sparsity ratio chosen for each network on each
dataset was the best achievable sparsity from
Table~\ref{tab:average_best_weights} minus 5\%.  This additional target allows
us to filter out very poor saliency metrics and to compare pruned networks of
similar size and accuracy.

The total cost of pruning is given by the sum of the cost of computing the
saliency metric and the cost of retraining.  The cost of a single retraining
step is the cost of one backward and one forward pass of the network. To easily
compare cost, we assume that the cost of a backward pass is twice that of a
forward pass.  The cost of computing a saliency metric is given in Table
\ref{tab:cost}.

\begin{table}[H]
\begin{center}
\caption{Cost of saliency metrics using $\mathcal{I}_{val}$ ($N_{val}$ batches of images).}
\label{tab:cost}
\scriptsize
\begin{tabular}{lllll}
\multicolumn{2}{l}{\textbf{Pointwise Metric}} & \textbf{Cost} & \textbf{Pointwise Metric} & \textbf{Cost}\\
\multirow{2}{*}{$x$}
 & $x=w$ & $0$ & $-x\diff{\loss}{x} + \frac{x^2}{2}\ddiff{\loss}{x}_{GN}$ & $3 \times N_{val}$\\
& $x=a$ & $1 \times N_{val}$ & $\frac{x^2}{2}\ddiff{\loss}{x}_{GN}$ & $3 \times N_{val}$\\
\multicolumn{2}{l}{$\diff{\loss}{x}$} & $3 \times N_{val}$ &$-x\diff{\loss}{x} + \frac{x^2}{2}\ddiff{\loss}{x}_{LM}$ & $5 \times N_{val}$\\
\multicolumn{2}{l}{$-x\diff{\loss}{x}$} & $3 \times N_{val}$ & $\frac{x^2}{2}\ddiff{\loss}{x}_{LM}$ & $5 \times N_{val}$\\
\end{tabular}
\end{center}
\end{table}

\begin{figure}
\centering
\includegraphics[width=\columnwidth]{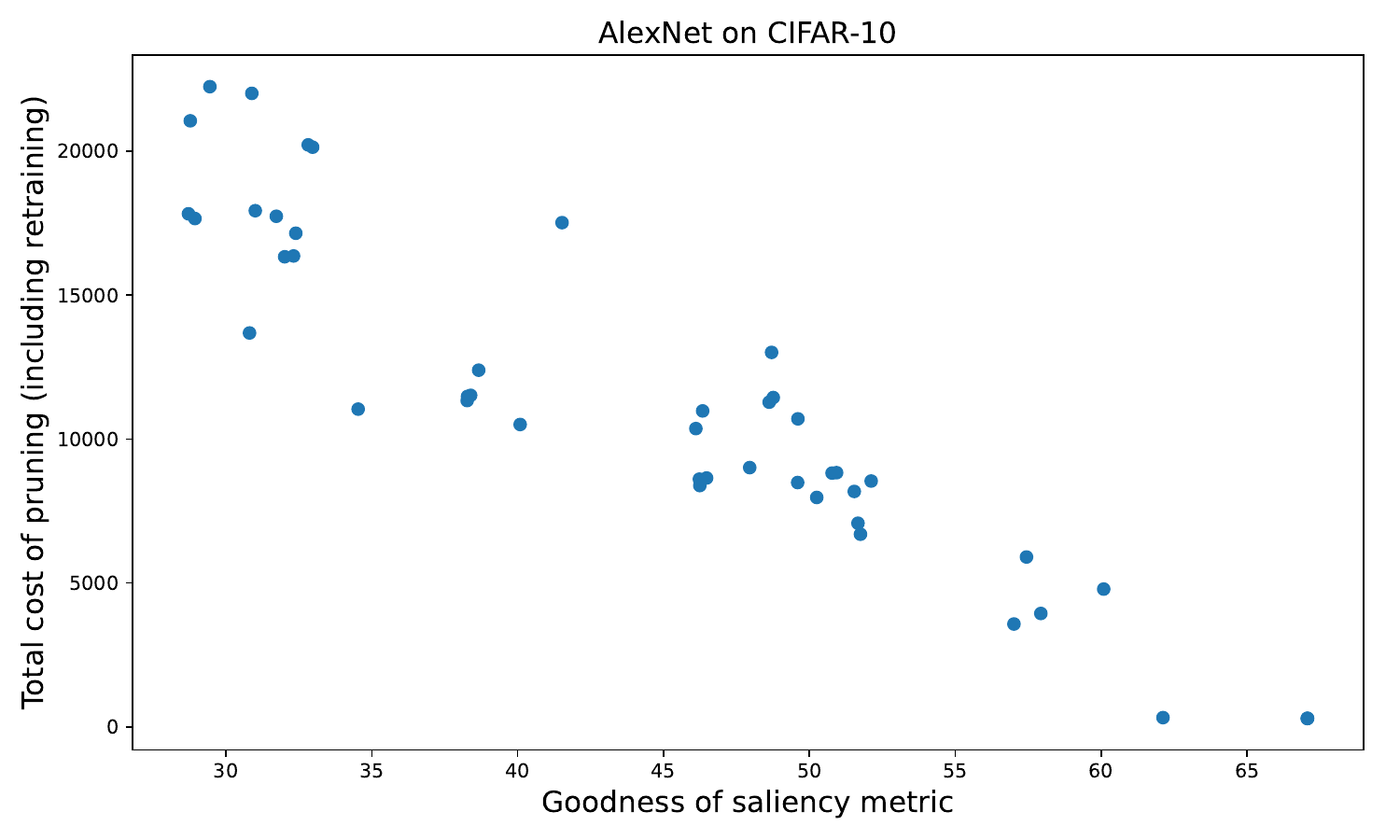}
\caption{Cost to prune a network (lower is better) against quality of the
saliency metric (higher is better).}
\label{fig:total_steps}
\end{figure}

In Figure \ref{fig:total_steps}, we see the result of this experiment for
AlexNet on the CIFAR-10 dataset. Each point on the graph is one saliency
metric. We see that many saliency metrics are able to be used in
Algorithm~\ref{alg:experiment} to meet both the sparsity and accuracy targets.
However, as predicted, poorer saliency metrics result in much more retraining
being required to reach a network of the given quality than good saliency
metrics. To quantify our results in terms of the correlation of metric quality
and pruning cost, we use the Spearman rank correlation. The correlation for
AlexNet on CIFAR-10 (as presented in Figure \ref{fig:total_steps} is $-0.9$. A
similar trend is observed (negative correlation in Table \ref{tab:correlation})
for the other networks except for ResNet-20 on CIFAR-10 where we had only 3
metrics meeting the targets.  In the case of AlexNet trained on CIFAR-10
(Figure~\ref{fig:total_steps}), with a fixed sparsity and accuracy target and
starting from the same initial trained network, the difference in total cost of
pruning was approximately 75$\times$ between the best and worst saliency
metric that meant both the sparsity and accuracy target.

\begin{finding} \emph{Better saliency metrics greatly reduce retraining requirements in
pruning.}
\end{finding}

\section{Conclusion}

Although many pruning strategies have been proposed, a common element
is the \textit{saliency metric} that seeks to identify unimportant
parameters. We propose a taxonomy that characterizes existing saliency
metrics, by combining elements from four mostly-orthogonal
components: (1) base input, (2) pointwise metric, (3) reduction,
and (4) scaling. This allows us to identify common components among saliency
metrics in the many existing pruning strategies, and to derive novel
metrics by combining elements from four mostly-orthogonal
components. We experimentally evaluate 308 such metrics.

We confirm some well-known results, like that gradient-based methods are
significantly better than simpler methods based purely on the weights or output
activations. But we also find new insights. For example, metrics that use the
gradient with respect to outputs tend to outperform those using the gradient
with respect to weights.  The most successful gradient-based methods use a
second-order Taylor expansion. Within this expansion, the first order term
contains important information that should not be omitted, but the Gauss-Newton
approximation is sufficient for the second order term.  Otherwise good metrics
can be easily undermined by a poor reduction, such as simply adding pointwise
terms.  On the other hand there is scope for significant improvements in
scaling factors containing structural information, such as our novel scaling
based on transitively-pruned parameters.

The saliency metric is just one component of pruning algorithms, but
it has a critical impact on the success of pruning. We anticipate that our
taxonomy and evaluation will guide practitioners to the best existing
saliency metrics, and direct researchers to open new frontiers in
the design space.

\section*{Acknowledgement}

This work was supported with the financial support of the Science Foundation
Ireland grant.  This work was also supported, in part, by Arm Research.


\printbibliography

\end{document}